%% file: main.tex
\pdfoutput=1
\documentclass[10pt,logo,copyright]{nvidiatechreport}
\usepackage[authoryear,sort&compress,square]{natbib}
\usepackage[frozencache,cachedir=.]{minted}

\usepackage{hyperref}
\usepackage{url}
\usepackage{graphicx}
\usepackage{booktabs} 
\usepackage{multirow}
\usepackage{multicol}
\usepackage{xspace}
\usepackage{algorithm}
\usepackage{dsfont}
\usepackage{algpseudocode}
\usepackage{algorithmicx}
\usepackage{subcaption}
\input{math_commands}

\input{header/macro}

\usepackage{wrapfig}

\usepackage[nameinlink]{cleveref}
\crefname{equation}{Eq.}{Eqs.}
\crefname{figure}{Fig.}{Figs.}
\crefname{section}{Sec.}{Sec.}
\crefname{appendix}{App.}{App.}
\crefname{table}{Tab.}{Tabs.}
\crefname{algorithm}{Alg.}{Alg.}
\crefname{thm}{Thm}{Thm}
\Crefname{thm}{Thm}{Thm}
\crefname{prop}{Prop}{Prop}
\crefname{line}{Line}{Lines}
\usepackage{longtable}

\let\cite\citep  

\newcommand{\crefnames}[3]{%
  \@for\next:=#1\do{%
    \expandafter\crefname\expandafter{\next}{#2}{#3}%
  }%
}

\title{Quantile Rendering: Efficiently Embedding High-dimensional Feature on 3D Gaussian Splatting}

\input{header/author}

\input{section/0_abstract}

\begin{document}
\maketitle
\abscontent

\input{section/1_introduction}

\input{section/2_related_work}
\input{section/3_preliminary}
\input{section/4_1_method}
\input{section/4_2_method}

\input{section/4_3_method}
\input{section/5_experiments}

\input{section/6_conclusion}

\appendix
\section*{Appendix}
\input{appendix/a_up_to_scale}
\input{figure/a_scene_scale/scene_scale}
\input{appendix/b_data_preprocessing}
\input{figure/b_gt_label/gt_label}
\input{appendix/c_theory}
\input{appendix/d_implementation_details}
\input{appendix/e_additional_exp}
\input{appendix/f_pseudocode}


\clearpage
\setcitestyle{numbers}
\bibliographystyle{plainnat}
\bibliography{iclr2026_conference}

\end{document}

%% file: math_commands.tex
\usepackage{amsmath,amsfonts,bm}

\def\Tableref#1{Table~\ref{#1}}
\def\lineref#1{line~\ref{#1}}

\def\Figref#1{Figure~\ref{#1}}

\def\secref#1{section~\ref{#1}}
\def\Secref#1{Section~\ref{#1}}

\def\eqref#1{eq.~\ref{#1}}

\def\Algref#1{Algorithm~\ref{#1}}

\def\1{\bm{1}}

\DeclareMathAlphabet{\mathsfit}{\encodingdefault}{\sfdefault}{m}{sl}
\SetMathAlphabet{\mathsfit}{bold}{\encodingdefault}{\sfdefault}{bx}{n}



%% file: header/macro.tex
\newcommand{\fullname}{Gaussian Splatting Network\xspace}
\newcommand{\nickname}{GS-Net\xspace}

\newcommand{\samplername}{Quantile Rendering\xspace}
\newcommand{\samplernameshort}{Q-Render\xspace}

\newcommand{\revision}[1]{#1}  

\newcommand{\set}[1]{ \{ #1 \} }
\newcommand{\noindentbold}[1]{\noindent\textbf{#1.}}

\newcommand{\GaussianFunction}{\mathrm{G}}
\newcommand{\bmu}{\mbox{\boldmath $\mu$}}
\newcommand{\Covariance}{\mathbf{\Sigma}}
\newcommand{\sph}{\mathbf{sph}}
\newcommand{\rot}{\mathbf{r}}
\newcommand{\Rot}{\mathbf{R}}
\newcommand{\scale}{\mathbf{s}}
\newcommand{\Scale}{\mathbf{S}}
\newcommand{\RGB}{\text{rgb}}

\newcommand{\feature}{\mathbf{f}}
\newcommand{\Feature}{\mathcal{F}}

\newcommand{\trans}{^{\top}}

\newcommand{\Real}{\mathbb{R}}

\newcommand{\Realpos}{\mathbb{R}_{>0}}

\newcommand{\Gaussian}{\mathcal{G}}
\newcommand{\gaussian}{\mathbf{g}}
\newcommand{\Quantile}{^{Q}}
\newcommand{\pixelcolor}{\mathbf{c}}
\newcommand{\ImageColor}{\mathbf{C}}

\newcommand{\ray}{\overrightarrow{p}}

\makeatletter
\DeclareRobustCommand\onedot{\futurelet\@let@token\@onedot}
\def\@onedot{\ifx\@let@token.\else.\null\fi\xspace}
\def\eg{\emph{e.g}\onedot} 
\def\ie{\emph{i.e}\onedot} 
 
\def\etc{\emph{etc}\onedot}

\def\aka{\emph{a.k.a}\onedot}

\newcommand{\mIoU}{mIoU ($\uparrow$)}
\newcommand{\mAcc}{mAcc ($\uparrow$)}

\newcommand{\Break}{\State \textbf{break}}



%% file: header/author.tex
\author{
Yoonwoo Jeong$^{1,2,\dagger}$, Cheng Sun$^1$, Frank Wang$^1$, Minsu Cho$^2$, Jaesung Choe$^1$\\
$^1$NVIDIA, $^2$POSTECH\\
}

%% file: section/0_abstract.tex
\begin{abstract}
\revision{
Recent advancements in computer vision have successfully extended Open-vocabulary segmentation (OVS) to the 3D domain by leveraging 3D Gaussian Splatting (3D-GS). 
Despite this progress, efficiently rendering the high-dimensional features required for open-vocabulary queries poses a significant challenge. 
Existing methods employ codebooks or feature compression, causing information loss, thereby degrading segmentation quality.
To address this limitation, we introduce Quantile Rendering (Q-Render), a novel rendering strategy for 3D Gaussians that efficiently handles high-dimensional features while maintaining high fidelity. 
Unlike conventional volume rendering, which densely samples all 3D Gaussians intersecting each ray, Q-Render sparsely samples only those with dominant influence along the ray. 
By integrating Q-Render into a generalizable 3D neural network, we also propose Gaussian Splatting Network (GS-Net), which predicts Gaussian features in a generalizable manner. 
Extensive experiments on ScanNet and LeRF demonstrate that our framework outperforms state-of-the-art methods, while enabling real-time rendering with an approximate ${\sim}43.7\times$ speedup on 512-D feature maps.
Code will be made publicly available.
}
\end{abstract}

%% file: section/1_introduction.tex
\section{Introduction}
\label{sec:introduction}

3D Gaussian Splatting (3D-GS)~\citep{3dgs} has emerged as a powerful representation for the neural rendering task, offering an explicit set of 3D Gaussians combined with efficient splatting~\citep{ewa} and tile-based rasterization for real-time rendering. 
This pipeline delivers both high-quality reconstruction and real-time frame rates—capabilities that have quickly made it a foundation for numerous 3D vision and graphics applications. 
\revision{
Beyond photorealistic rendering, recent works~\citep{langsplat,opengaussian,gaussian_grouping,gaga,dr_splat} have begun leveraging 3D Gaussians as a medium for scene understanding.
A prevalent approach is to distill knowledge from powerful 2D foundation models—such as CLIP~\citep{clip,openclip}, SAM~\citep{sam,sam2}, and DINO~\citep{dino,dinov2,dinov3}—into pre-optimized 3D Gaussians by embedding high-dimensional features.
In this work, we focus on open-vocabulary segmentation (OVS) using OpenCLIP~\citep{openclip}, which outputs 512-dimensional language-aligned features for arbitrary text or image queries.
}

The original volume rendering algorithm~\citep{3dgs} samples \emph{all} Gaussians intersecting a ray, regardless of their actual contribution to the output,~\ie rendered pixel color. For RGB rendering this overhead is manageable, but for high-dimensional embeddings (e.g., 512-D CLIP features) it becomes computationally heavy. To resolve this problem, series of studies compress the dimensionality of 512-D CLIP features into 3-D or 6-D features or codebooks~\citep{langsplat,opengaussian,feature_gs,dr_splat}. 
While effective, this strategy is not a fundamental solution and can potentially loss the original information that was stored in the high-dimensional features. Moreover, the distribution of the optimized 3D Gaussians potentially have noisy or local minima due to its per-scene optimization scheme. Accordingly, it is challenging to properly embed high-dimensional feature vectors on the top of these 3D Gaussians.

Based on this observation, we hypothesize that not all Gaussians are influential—only a partial fraction of 3D Gaussians meaningfully affect the high-dimensional feature rendering along a ray. This observation motivates Quantile Rendering (Q-Render), our transmittance-aware and efficient rendering algorithm for high-dimensional Gaussian features. Instead of densely accumulating every rasterized 3D Gaussian, Q-Render adaptively selects a small set of \emph{quantile Gaussians}—those that dominate the ray’s transmittance profile—and renders only these representatives. This quantile-based selection cuts redundant computation, and approximates the original signals for downstream tasks that may require to render high-dimensional feature maps.

In this work, integrating Q-Render into a 3D neural network~\citep{mink,ptv3,splatformer}, we build \fullname (\nickname) that operates on 3D Gaussians to predict Gaussian features. 
Typically, Q-Render serves as an efficient bridge between 2D supervision and the 3D neural network, allowing backward gradients~\citep{backprop} to flow from image-space losses to the 3D neural network's predictions. Moreover, with this integration, Q-Render’s sparse sampling becomes even more advantageous: the inductive bias of the 3D neural network tends to predict spatially smooth Gaussian features, meaning that densely sampling all Gaussians along each ray is unnecessary. Instead, the sparsely selected quantile Gaussians are sufficient to faithfully render high-dimensional feature maps while significantly reducing computational overhead during the rendering process and its backward computation. 

We validate our method on two open-vocabulary 3D semantic segmentation benchmarks: (1) ScanNet and (2) LeRF-OVS, where CLIP-based embeddings are stored directly in 3D Gaussians. In both cases, Q-Render achieves state-of-the-art results, underscoring its value as a scalable bridge between 2D foundation models and 3D Gaussian representations.
\begin{itemize}
    \item \emph{Quantile Rendering} — a sparse, transmittance-guided sampling strategy that selects only the most representative Gaussians along each ray for efficiency.
    \item \emph{Gaussian Splatting Network} — A 3D neural network that predicts high-dimensional featuremetric Gaussians from optimized 3D Gaussians, with Q-Render enabling efficient and effective feature distillation from 2D foundation models.
    \item \emph{Extensive Validation} — comprehensive experiments on open-vocabulary 3D semantic segmentation benchmarks with superior performances against recent studies.
\end{itemize}

\input{figure/1_teaser/teaser}

%% file: figure/1_teaser/teaser.tex
\begin{figure}[t!]
    \vspace{-2mm}
    \centering
    \includegraphics[width=\linewidth]{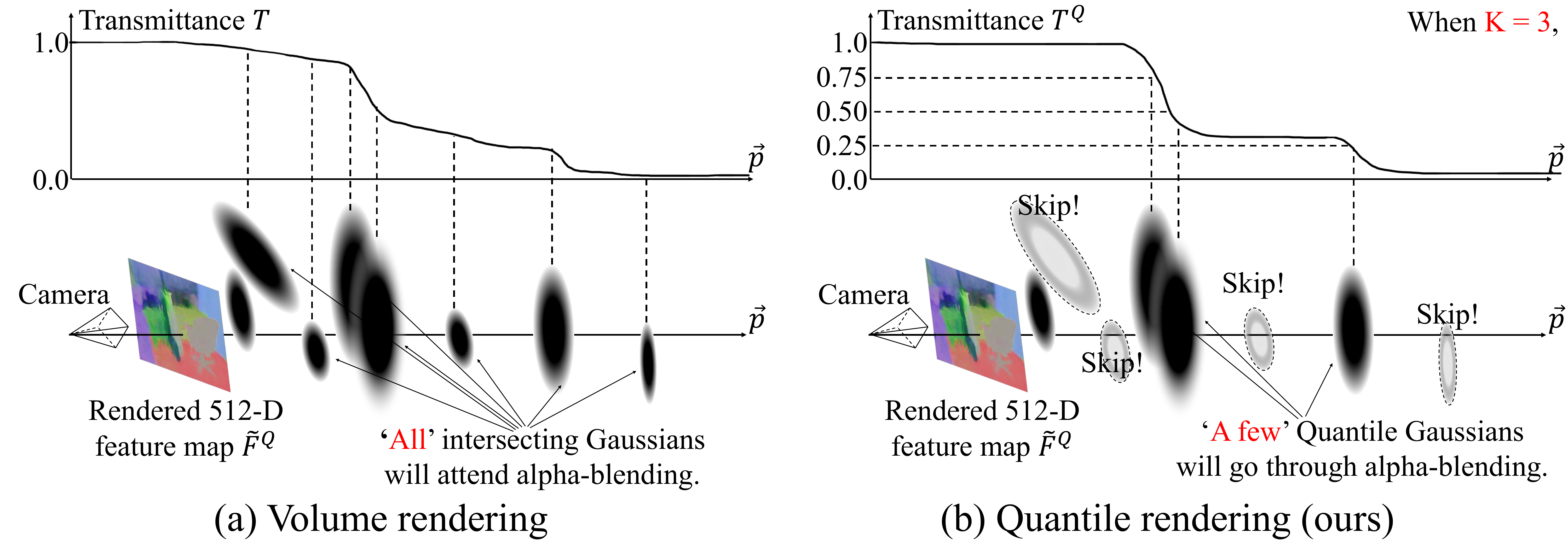}
    \caption{
        Quantile rendering. (a) Unlike volume rendering~\citep{nerf,3dgs} that densely samples and blends all 3D Gaussians along the rays, (b) our Quantile render selectively samples and blends a sparse set of quantile Gaussians -- those with dominant influence along the ray, which can efficiently render high-dimensional feature maps from Gaussian features.
    }
    \label{fig:teaser}
\end{figure}

%% file: section/2_related_work.tex
\section{Related works}
\label{sec:related_works}

\paragraph{High-dimensional Features on Gaussian representation.}
There is a growing body of this field that adopts 3D Gaussian Splatting~\citep{3dgs} as the underlying representation, incorporating these language-aligned embeddings into per-Gaussian memory~\citep{langsplat,legaussian,feature_gs,opengaussian,dr_splat}. 
The core idea is to utilize a rendering pipeline of vanilla 3DGS~\citep{3dgs} to sample 3D Gaussians for distillation from 2D image embeddings, a capability that point clouds inherently lack.
Nonetheless, their memory-based approaches force per-scene feature optimization, which limits their application in practice. Moreover, lifting high-dimensional embeddings (\eg, CLIP 512-D feature) into 3D Gaussians demands significant computational resources.


\paragraph{Neural scene representation networks.} 
Recent research has explored treating neural scene representations as 3D models, using neural networks to process them and solve various 3D tasks. PeRFception~\citep{perfception} trains networks to process Plenoxels~\citep{plenoxels} for classification and semantic segmentation. SPARF~\citep{sparf} improves Plenoxel in a few-shot setup by training networks that process few-shot Plenoxels and output the enhanced version. SplatFormer~\citep{splatformer} adopts Point Transformer V3~\citep{ptv3} to improve the robustness of 3D-GS under out-of-distribution poses. To the best of our knowledge, we are the first to address language and grouping tasks using networks that process 3D-GS. 

%% file: section/3_preliminary.tex
\section{Preliminary}
\label{sec:preliminary}


\paragraph{3D Gaussian Splatting.}
3D-GS~\citep{3dgs} represents 3D scenes as a set of 3D Gaussians where each 3D Gaussian~$\gaussian$ is parameterized as center position~$\bmu \in \Real^3$, covariance matrix~$\Covariance \in \Real^{3\times 3}$, opacity $\alpha \in \Real$, and spherical harmonics coefficients with $d$~degree $\sph \in \Real^{(d+1)^2\times 3}$. In particular, the covariance matrix of the anisotropic Gaussian distribution is decomposed into scaling factors $\scale \in \Realpos^3$, rotation in quaternion representation~$\rot \in \Real^4$, where $\Covariance = \Rot \Scale \Scale\trans \Rot\trans$.
Given a set of $N$ numbers of 3D Gaussians $\Gaussian = \set{\gaussian_i}_{i=1}^N = \set{\bmu_i, \scale_i, \rot_i, \alpha_i, \sph_i}_{i=1}^N$, 3D-GS~\citep{3dgs} organizes the rendering pipeline as: (1) tiling strategy; (2) splatting algorithm~\citep{ewa}; and (3) volume rendering~\citep{nerf}. 
In details, volume rendering proceeds with alpha-blending through densely rasterized 3D Gaussians along a ray~$\ray$ as:
\vspace{2mm}
\begin{equation}\label{eq:volumetric rendering} 
    \Tilde{\ImageColor}[\ray] = \sum_{i\in S_r}\! w_i\pixelcolor_i~~~\text{s.t.}~~~w_i = T_i\alpha'_i,~~T_i = \!\!\!\!\!\! \prod_{j \in S_{r, :i-1}} \!\!\!\!\! (1-\alpha'_j),~~\alpha'_i = \alpha_i \cdot \GaussianFunction_{\bmu_i, \Covariance_i}(u)
\end{equation}
where $\Tilde{\ImageColor}$ is a rendered image, $\Tilde{\ImageColor}[\ray]$ is a rendered pixel color at the ray~$\ray$, $\pixelcolor_i$ is the emitted color of the $i$-th Gaussians obtained from $\sph_i$, $S_r$ is an ordered sequence of indices of densely sampled Gaussians along $\ray$, $T_i$ is the transmittance at $i$-th Gaussian along $\ray$, and $\GaussianFunction(\cdot)$ is the Gaussian function evaluated at pixel location $u$ by projecting the Gaussian parameters.
%
For training, 3D-GS optimizes parameters~$\Gaussian$ by minimizing a rendering loss~$\mathcal{L}_\RGB = \| \ImageColor - \Tilde{\ImageColor} \|_1 + \lambda_{\text{SSIM}}\cdot \operatorname{SSIM}(\ImageColor, \Tilde{\ImageColor})$ where $\ImageColor$ is a ground truth image and $\lambda_{\text{SSIM}}$ is the hyperparameter adjusting the SSIM loss.

\paragraph{High-dimensional Gaussian features.}
Recent studies~\citep{langsplat,opengaussian,gaussian_grouping,gaga,feature_gs} propose to register high-dimensional features for each 3D Gaussian~$\gaussian$. These embedded features store 3D scene information such as language attributes, mask identities,~\etc.
These methods commonly start from optimized 3D Gaussians~$\Gaussian$ following the original 3D-GS paper~\citep{3dgs}. Then, while freezing~$\Gaussian$, this method allocates $C$-dimensional feature vector~$\feature \in \Real^{C}$ for every 3D Gaussian~$\gaussian$ such that these feature vectors are optimized to minimize feature rendering losses.


%% file: section/4_1_method.tex
\section{Methodology}
\label{sec:methodology}

\input{figure/2_overview/overview}

As shown in~\Figref{fig:overview}, given a set of $N$ numbers of optimized 3D Gaussians~$\Gaussian = \{ \gaussian_i \}_{i=1}^N$, a 3D neural network predicts a set of $C$-dimensional Gaussian features~$\Feature = \{ \feature_i \}_{i=1}^N$ where $\feature \in \Real^{C}$. 
Through our Quantile Rendering, the Gaussian features are rendered into $C$-dimensional feature maps. The 3D neural network is trained to minimize the rendering loss between the rendered feature maps and the target 2D feature maps extracted from CLIP's vision encoder.

\subsection{3D neural network}
\label{subsec:3d_neural_network}
Conventional 3D neural networks for pointcloud~\citep{mink,ptv3,choe2021deep,regionplc,pla,mosaic3d} often voxelize 3D points to efficiently process scene-scale 3D points. They transform scattered points into sparse voxel grids with unique spatial locations, enabling single-pass inference over an entire 3D scene thanks to its efficiency. In contrast to handling pointclouds, modeling a scene with continuous 3D Gaussians~\citep{3dgs} introduces overlapping regions since each Gaussian has a volumetric extent. To resolve this issue, we follow SplatFormer~\citep{splatformer} that introduces voxelization on 3D Gaussians by sampling its center location~$\bmu$ to ensure compatibility with typical 3D backbones: Point Transformer v3 (PTv3)~\citep{ptv3} and MinkUnet~\citep{mink}. These models are designed to predict voxel features such that we proceed with the de-voxelization steps to convert predicted voxel features into predicted Gaussian features~$\Feature$, which will be used for rendering procedures.

%% file: figure/2_overview/overview.tex
\begin{figure*}[t!]
    \centering
    \includegraphics[width=1.0\linewidth]{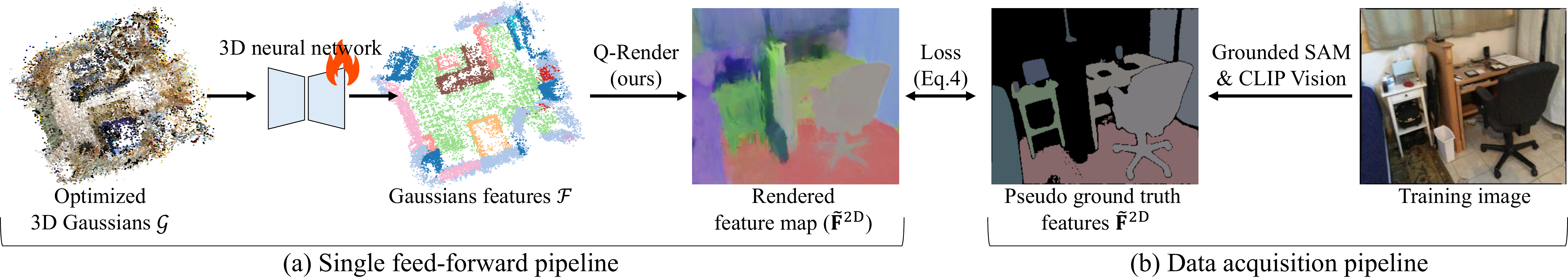}
    \caption{Overview. Given optimized 3D Gaussians~$\Gaussian$, our network is trained to predict Gaussian features~$\Feature$ that are aligned with the language embedding space from CLIP's vision encoder. Typically, the proposed Q-Render accelerates the training and inference speed by transforming predicted Gaussian features into rendered feature maps.
    }
    \label{fig:overview}
\end{figure*}

%% file: section/4_2_method.tex
\subsection{Quantile rendering}
\label{subsec:quantile}

To train the 3D neural network using the knowledge from 2D foundation models, rendering process is necessary. Volume rendering~\citep{nerf,3dgs} can be an option, but it may require large computation power~\citep{langsplat,opengaussian,feature_gs} when rendering high-dimensional Gaussian features. This is because volume rendering repeatedly accumulates high-dimensional Gaussian features~$\Feature$ along the ray, which quickly becomes prohibitively expensive as dimensionality grows as described in~\Tableref{table.complexity}.

To address this, we introduce a quantile-based sampling strategy. Quantiles---statistical cutpoints that divide a distribution into intervals of equal probability---enable sparse yet representative sampling. By approximating the distribution in traditional volume rendering with a carefully selected subset of Gaussians, our \samplername achieves efficient rendering while preserving representativeness. Specifically, given a hyperparameter $K$, the algorithm samples $K$ Quantile Gaussians along each ray. The detailed procedure is presented in~\Algref{alg:quantile_based_rendering}. 
We first rasterize and splat 3D Gaussians following~\citep{3dgs} to obtain the indices of rasterized 3D Gaussians~$I$ at the target ray~$\ray$ as described in~\Secref{sec:preliminary}. Then, our Q-Render takes place to run three sub-steps: Quantile Gaussian sampling, alpha-blending, and feature normalization.



\input{algorithm/gaussian_sampling}

\noindentbold{Sampling Quantile Gaussian}
Given hyperparameter $K$, we partition the transmittance~$T\in\Real_{[0,1]}$ into $K{+}1$ evenly distributed segments at each ray, and track how much the transmittance changes by passing through each Gaussian.
Once crossing a segment boundary\revision{(}~\lineref{line:transmittance_check}~of~\Algref{alg:quantile_based_rendering}\revision{)}, we determine that this Gaussian as Quantile in this interval.

\noindentbold{Alpha-blending on Quantile Gaussians} 
With these Quantile Gaussians that show the relatively big transmittance change, we participate such Gaussians in alpha-blending. As summarized in~\Tableref{table.complexity}, this selective blending reduces the time complexity from $\mathcal{O}(NC)$ (volume rendering) to $\mathcal{O}(N + KC)$, where $N$ is the number of Gaussians for each pixel and $C$ is the feature dimension. 
Unlike top-$K$ sampling proposed by a concurrent study~\citep{dr_splat}, which requires additional complexity to sort values, $\mathcal{O}(N\log K + KC)$.

\noindentbold{Normalizing feature vector}
In the volume rendering~\citep{3dgs,nerf}, transmittance~$T$ is initialized as one and then the remaining transmittance~$T$ closes to zero as proceeding with the alpha blending along a ray.
However, alpha-blending on sub-sampled Gaussians may have remaining transmittance that has relatively higher than 0. To approximate the original distribution in volume rendering that returns zero-close final transmittance value, we forcefully set the final transmittance~$T\Quantile$ as zero by normalizing the accumulated feature $\Tilde{\feature}\Quantile$ as~$\Tilde{\feature}\Quantile \gets \frac{\feature\Quantile}{1 - T\Quantile}$.

\input{table/complexity}
\input{figure/3_transmittance/transmittance}

Finally, our Quantile rendering has been formulated as $\Tilde{\feature}\Quantile = \operatorname{QuantileRender}(\Gaussian, \Feature, K, I)$
where $I$ is the indices of rasterized 3D Gaussians.
As illustrated in~\Figref{fig:quantile}, \samplernameshort well approximates the transmittance tendency from the volume rendering~\citep{3dgs} while top-$K$ sampling strategy~\citep{dr_splat} shows different the tendency. In short, our method avoids the overhead and outperforms top-$K$ in both efficiency and accuracy as stated in~\Tableref{fig:render speed}.

\revision{
Furthermore, the inductive bias of the 3D neural network promotes spatially smooth Gaussian feature predictions. Accordingly rendering dense sampling along rays becomes redundant. Our sparse quantile selection remains sufficient for high-fidelity feature mapping while significantly reducing computational overhead during both rendering and backward passes. Such an approximation is bounded by $O(1/K)$ where we provide a detailed theoretical justification of Q-Render as a Riemann sum approximation in Appendix C.
}

%% file: algorithm/gaussian_sampling.tex
\begin{algorithm}[t]
    \caption{\samplername}
    \label{alg:quantile_based_rendering}
    \begin{algorithmic}[1]
    \Require Optimized 3D Gaussians~$\Gaussian$, predicted Gaussian features~$\Feature$, the number of Quantile Gaussians~$K$, indices of rasterized 3D Gaussians~$I$ at target ray.
    \Ensure Rendered feature vector at the target ray~$\Tilde{\feature}\Quantile \in \Real^C$.
    \Procedure{QuantileRender}{$\Gaussian$,~$\Feature$,~$K$,~$I$}
        \State $T\gets 1$,~~$T\Quantile \gets 1$,~~$\feature\Quantile \gets \mathbf{0}$,~~$k \gets 0$
        \For{$i$ in $I$}
            \State $\gaussian_i \gets \Gaussian[i]$, $\feature_i \gets \Feature[i]$  \Comment{$\gaussian_i = \{ \bmu_i, \scale_i, \rot_i, \alpha_i, \sph_i \}$}
            \State $T_\text{test}$ $\gets$ $T\cdot(1 - \alpha'_i)$
            \If{$T_{\text{test}} < 1 - \frac{k+1}{K+1}$} \label{line:transmittance_check}  \Comment{Sampling Quantile Gaussian}
                \State $k \gets$ $k+1$
                \State $w\Quantile$ $\gets$ $T\Quantile\cdot \alpha'_i$  
                \State $\feature\Quantile$ $\gets$ $\feature\Quantile + w\Quantile \cdot \feature_i$  \Comment{Alpha-blending on Quantile Gaussians}
                \State $T\Quantile$ $\gets$ $T\Quantile \cdot (1-\alpha'_i)$
                \While{$T_{\text{test}} < 1 -  \frac{k+1}{K+1}$}
                \State $k$ $\gets$ $k+1$
                \EndWhile
            \EndIf
            \If{$T_{\text{test}} < \frac{1}{K+1}$}
                \Break
            \EndIf
            \State $T$ $\gets$ $T_\text{test}$
        \EndFor
        \State $\Tilde{\feature}\Quantile$ $\gets$ $\frac{\feature\Quantile}{1 - T\Quantile}$ \Comment{Normalizing feature vector}
        \State \Return $\Tilde{\feature}\Quantile$
    \EndProcedure
    \end{algorithmic}
\end{algorithm}

%% file: table/complexity.tex
\begin{table}[t]
    \centering
    \caption{
        Complexity comparison table. $K$ intervals along each ray, $N$ is the number of Gaussians per pixel and $C$ is the feature dimension. It shows the efficiency of our \samplernameshort, which avoids any multiplier for the large $N$. Precise inference time is in~\Figref{fig:render speed}.
    }
    \resizebox{0.95\linewidth}{!}{
    \begin{tabular}{c|ccc}
    \toprule
    
    & V-Render~\citep{3dgs} & top-$K$~\citep{dr_splat} & \samplernameshort (ours) \\ 
    \midrule
    
    Complexity & $\mathcal{O}(NC)$ & $\mathcal{O}(N\log K + KC)$ & $\mathcal{O}(N + KC)$ \\ 
    \bottomrule
    \end{tabular}}
    \vspace{4mm}
    \label{table.complexity}
\end{table}

%% file: figure/3_transmittance/transmittance.tex
\begin{figure}[t!]
    \centering
    \includegraphics[width=0.65\linewidth]{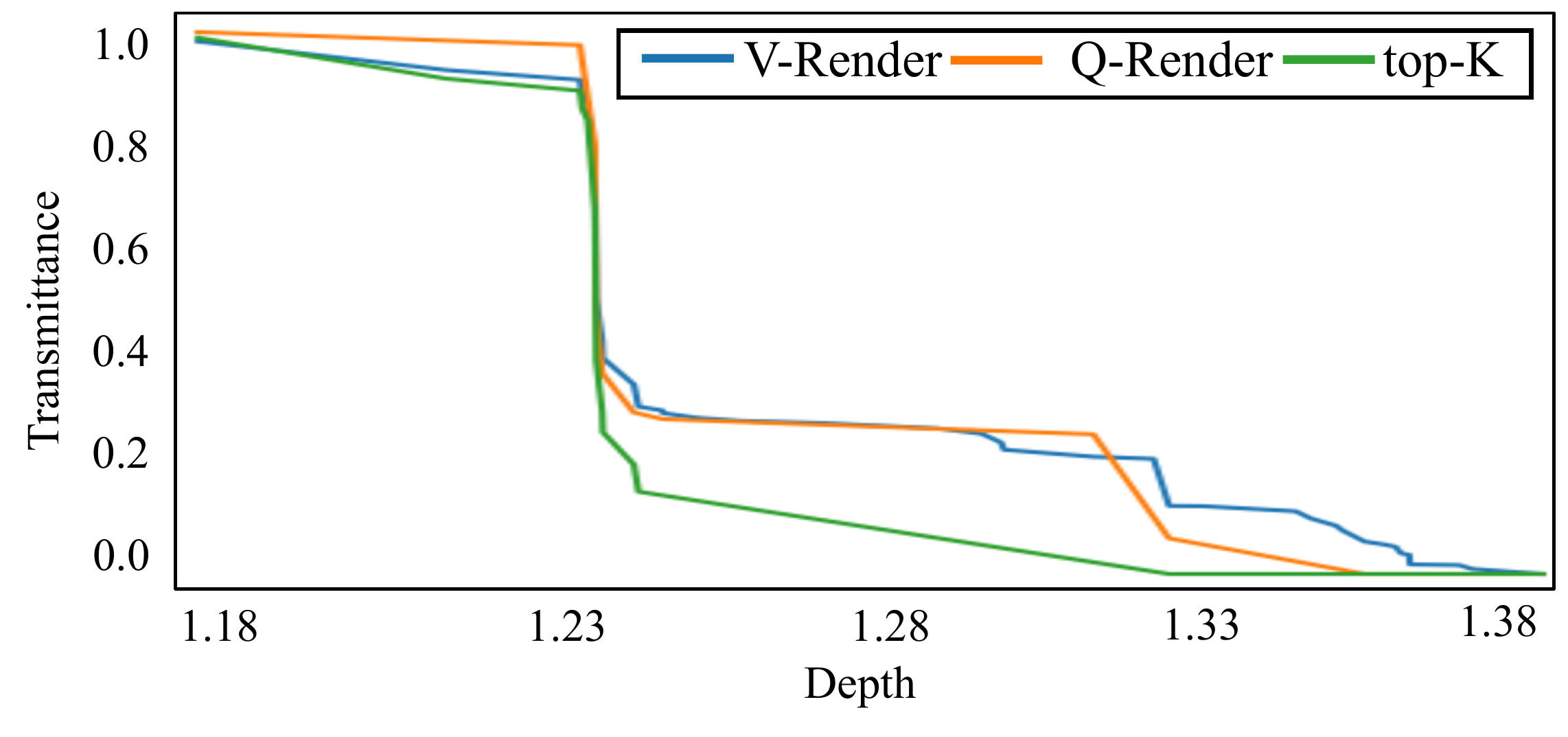}
    \vspace{-2mm}
    \caption{Comparison of transmittance distribution across different Gaussians sampling algorithms. Our \samplernameshort effectively approximates the distribution of the transmittance distribution of the original 3D-GS. We used $K=10$ for visualization.}
    \label{fig:quantile}
\end{figure}

%% file: section/4_3_method.tex
\subsection{Training loss}
\label{subsec:training_loss}

We validate our \nickname in open-vocabulary 3D semantic segmentation task with two datasets, ScanNet dataset and LeRF-OVS dataset. In this task, CLIP~\citep{clip,openclip} is one of the popular vision language models that predicts a 512-D feature vector from a given input image. Following the previous studies~\citep{langsplat,opengaussian,dr_splat}, we set the distillation target as CLIP vision encoder's feature vectors. We first extract image patches,~\aka masks~$\{ \mathbf{m}\}$, using Grounded-SAM2~\citep{groundedsam}, and extract CLIP embeddings from the corresponding patch~$\feature^\text{CLIP}$.
Given a pair of training data $\{ \mathbf{m}_i, \feature^\text{CLIP}_i \}$, the 3D neural network is trained to minimize a discrepancy between our rendered feature vector~$\Tilde{\feature}\Quantile$ and CLIP embedding~$\feature^\text{CLIP}$ in the format of the contrastive loss~$\mathcal{L}$ as follows:
\begin{equation}\label{eq:ovs loss}
    \mathcal{L} = -\log \frac{
        \exp(\operatorname{sim}(\Tilde{\feature}\Quantile,~\feature^\text{CLIP}_i))
    }{
        \sum_{i \neq j} \exp(\operatorname{sim}(\Tilde{\feature}\Quantile,~\feature^\text{CLIP}_j))
    },
\end{equation}
where $\operatorname{sim}(\cdot,\cdot)$ is the cosine similarity function.

%% file: section/5_experiments.tex
\section{Experiments}
\label{sec:experiments}

\input{table/openvocab.scannet}
\input{figure/5_scannet_qual/scannet_qual}

We provide experiments to verify the efficacy of our \nickname on open-vocabulary 3D semantic segmentation task using the ScanNet dataset and LeRF-OVS dataset. Given a set of open-vocabulary text queries, we extract their text features using CLIP's text encoder. For fair comparison, we used the sam CLIP model provided by the recent work~\citep{dr_splat}. We then compute the cosine similarity between Gaussian features and these text features, assigning each Gaussian the labels with the highest similarity scores.

Following the evaluation protocol of OpenGaussian~\citep{3dgs}, we assess mIoU and mAcc on predefined sets of 19, 15, and 10 categories by theirs. In contrast to OpenGaussian, which freezes the original pointclouds and disables densification and pruning, we enable both densification and pruning to fully leverage the capacity of 3D-GS. As a result, per-Gaussian labels are required for evaluation that is extracted from pointcloud labels. 
To assign semantic labels to each Gaussian, we use the Mahalanobis distance to find the \( K \) nearest neighbors, similar to Dr.Splat~\citep{dr_splat}. Furthermore, we found that filtering points based on opacity and Gaussian influence leads to more reliable labels, as reported in~\secref{subsec:Pseudo ground truth acquisition}. For a fair comparison, we reproduce the results of LangSplat~\citep{langsplat}, OpenGaussian~\citep{opengaussian}, and Dr.Splat~\citep{dr_splat} using our training and evaluation setup. 

\subsection{ScanNet dataset}
\label{subsec:scannet_dataset}
We compare GS-Net against baselines~\citep{langsplat, opengaussian} on the ScanNetv2~\citep{scannet} dataset, which includes 1,513 scenes and 124,505 frames of indoor captures featuring various household objects. When training our generalized network, we use the same 10 validation scenes as OpenGaussian~\citep{opengaussian} and train on the remaining 1,503 scenes. Additionally, we train GS-Net to overfit a single scene to ensure a fair comparison with the baselines. We pre-optimize all 3D-GS~\citep{3dgs} using the original implementation across all scenes to train and evaluate \nickname.  

In Table~\ref{tab:openvocab.quant}, we compare two \nickname models, GS-Mink and GS-PTv3 where each model has MinkUnet~\citep{mink} and PTv3~\citep{ptv3} as baseline 3D neural networks, against previous baselines~\citep{langsplat, opengaussian, dr_splat}.
Although the evaluation scenes are not used for training \nickname, \nickname achieves significant performance improvements or comparable results to previous baselines.
Furthermore, when overfitted to a single scene, GS-Mink and GS-PTv3 improve mIoU by $12.08\%p$ and $12.73\%p$, respectively.
Interestingly, while GS-PTv3 outperforms GS-Mink in the overfitting scenario, this trend reverses when training for generalization.
We found that GS-PTv3 is more prone to overfitting during training, suggesting that architectural improvements for handling Gaussians could mitigate this issue.
As shown in Figure~\ref{fig:openvocab.qual}, our \nickname produces much clearer semantic segmentation results, further demonstrating the superiority of our method.

\input{table/openvocab.lerf_ovs}
\input{figure/6_lerf_ovs_qual/lerf_ovs_qual}

\subsection{LeRF-OVS dataset.} 
\label{subsec:lerf_ovs_dataset}
We compare GS-Mink with previous baselines on LeRF-OVS dataset~\citep{lerf,langsplat} where the scenes has sampled for the open vocabulary semantic segmentation task such `ramen', `teatime', `kitchen', \etc. we maintain to use the same ground truth masks and corresponding text captions for the fair comparison. We use GS-Mink as our baselines and train the model with this dataset. As demonstrated in~\Tableref{table:lerf_dataset}, our method outperforms recent studies while having 6-D compressed Gaussian embeddings and 512-D Gaussian embeddings as the same dimensionality as CLIP's image embeddings. For visualization, we also provide our qualitative results in~\Figref{rebutal.fig.rendered_featmap}.

\input{figure/4_fps_miou/fps_miou}

\subsection{Control experiments}
\label{subsec:control_experiments}

\noindentbold{Feature renderer} We conduct ablation study by replacing our Q-Render of our GS-Net with other feature rendering algorithms, such as volume render~\citep{3dgs}, and the top-$K$ render~\citep{dr_splat}. The comparison results are described in~\Figref{fig:render speed} in terms of rendering speed and accuracy in open-vocabulary semantic segmentation.

In general, \samplernameshort achieves superior or comparable performance compared to other feature rendering algorithms. In terms of rendering speed, our Q-Render demonstrate up to $1.5\times$ faster speed in comparison with volume rendering (V-Render). Moreover, top-$K$ rendering shows remarkable speed drops as $K$ increases. Such a phenomenon is related to our complexity analysis in~\Tableref{table.complexity}.

For the detailed ablation study about $K$, the number of quantile Gaussians, we achieve the best performance when setting $K=40$. Nonetheless, the performance becomes converged as $K\ge10$, and reaches $\sim50$mIoU. Furthermore, \Tableref{table.complexity} shows that top-$K$ render has relatively big performance drop when setting smaller $K$. We deduce that this is because of the discrepancy in the transmittance profile as we visualized in \Figref{fig:quantile}.

In another analysis, our Q-render with $K=40$ outperforms the performance of the volume render. Theoretically speaking, our Q-render is designed to approximate the original transmittance profile by the volume render as stated in~\secref{subsec:quantile}. Though we do not have concrete experimental supports, we guess that this is related to the potential noise in the optimized 3D Gaussians~$\Gaussian$ which have some difficulties in representing the precise geometry information due to the limited training images~\citep{zhu2024fsgs} or the 3D Gaussian representation itself~\citep{2dgs}.

\noindentbold{Grid size}
Throughout the paper, we used a grid size of $10.0$cm for both GS-Mink and GS-PTv3. We also conducted ablation studies by varying the grid size to $10.0$, $5.0$, $2.0$, $1.0$, $0.50$, and $0.25$ cm.
The results of open vocabulary semantic segmentation experiments with different grid sizes are presented in~\Tableref{tab.grid_size}. We observed that reducing the grid size to $5.0$cm shows the best performance. While the performance started to dramatically drop after setting grid size smaller than $1.0$ cm. We deduce this phenomenon from the model architecture and voxelization strategy. While we follow the Gaussian voxelization scheme by~\citep{splatformer}, it only sample one voxel per Gaussian though each Gaussian has volumetric shapes. Moreover, the 3D neural network baselines have some receptive field extent. Once we reduce the grid size smaller than specific numbers, some voxels may not aggregated together due to the receptive field limitations. Based on this analysis, we believe that exploring model designs or voxelization type can become potential future research directions for the further improvements.

\input{table/ablation.grid_size}
\input{table/rebuttal.inference_time}

\noindentbold{Inference time} 
In~\Tableref{rebuttal.table.inference_time}, we evaluate the inference speed of recent methods and our GS-Mink on scene0000\_00 from ScanNet. For a fair comparison, we measure frames per second (FPS) using the same feature dimension settings as in LangSplat ($3$ dim) and OpenGaussian ($6$ dim), and utilize the full Gaussian scene. Our method achieves the highest rendering speed among the evaluated approaches.
We modify the official implementation by \citep{langsplat} and \citep{opengaussian} to render 512-D feature maps by for-loop iterations (we denoted these implementations as 512\textsuperscript{\textdagger} in~\Tableref{rebuttal.table.inference_time}). It turns out that our Q-Render achieves upto ${\sim}\textbf{43.7}\times$ speed gains when rendering 512-D feature maps. 


\noindentbold{Information loss after voxelization} 
We visualize the voxelization results in~\Figref{rebuttal.fig.info_loss}. The voxelization itself potentially brings information loss, and changes the original 3D Gaussian parameters, such as opacity, spherical harmonics, \etc. To estimate the amount of information loss, we conduct an experiment that renders images from (1) original Gaussians, or (2) voxelized \& de-voxelized Gaussians. \Figref{rebuttal.fig.info_loss} shows that PSNR drops from $19.89$ to $15.19$ when $\delta{=}0.10$. Due to these reasons, we de-voxelize the estimated voxel features from the 3D neural network into Gaussian features. Then, we proceed with our Q-rendering. We used \citep{svraster} to directly render images from sparse voxels without de-voxelization steps.

\input{figure/7_info_loss/info_loss}

%% file: table/openvocab.scannet.tex
\begin{table*}[!tb]
    \caption{
        Open vocabulary 3D semantic segmentation performances in the ScanNet dataset. 
    }
    \centering
    \resizebox{1.0\linewidth}{!}{
    \begin{tabular}{l|c|cc|cc|cc}
    \toprule
    \multicolumn{1}{c|}{\multirow{2}{*}{Method}} & \multicolumn{1}{c|}{Per-scene} & \multicolumn{2}{c|}{19 classes} & \multicolumn{2}{c|}{15 classes} & \multicolumn{2}{c}{10 classes} \\
    
     & optim. & \mIoU & \mAcc & \mIoU & \mAcc & \mIoU & \mAcc \\
    \midrule
    
    LangSplat~\citep{langsplat} & \checkmark & 1.47 & 10.23 & 2.00 & 11.85 & 3.24 & 16.79 \\
    OpenGaussian~\citep{opengaussian} & \checkmark & 22.60 & 34.41 & 24.21 & 37.58 & 34.74 & 51.58 \\
    \midrule
    
    Dr.Splat~\citep{dr_splat} & & 23.21 & 35.42 & 25.33 & 34.64 & 36.71 & 53.29 \\
    GS-Mink (ours) & & \textbf{50.75} & \textbf{62.00} & \textbf{53.54} & \textbf{66.39} & \textbf{64.95} & \textbf{79.34}  \\
    GS-PTv3 (ours) & & 48.99 & 60.36 & 52.39 & 66.05 & 62.57 & 77.70  \\
    \bottomrule
    \end{tabular}}
    \label{tab:openvocab.quant}
\end{table*}

%% file: figure/5_scannet_qual/scannet_qual.tex
\begin{figure*}[t!]
    \centering
    \includegraphics[width=1.0\linewidth]{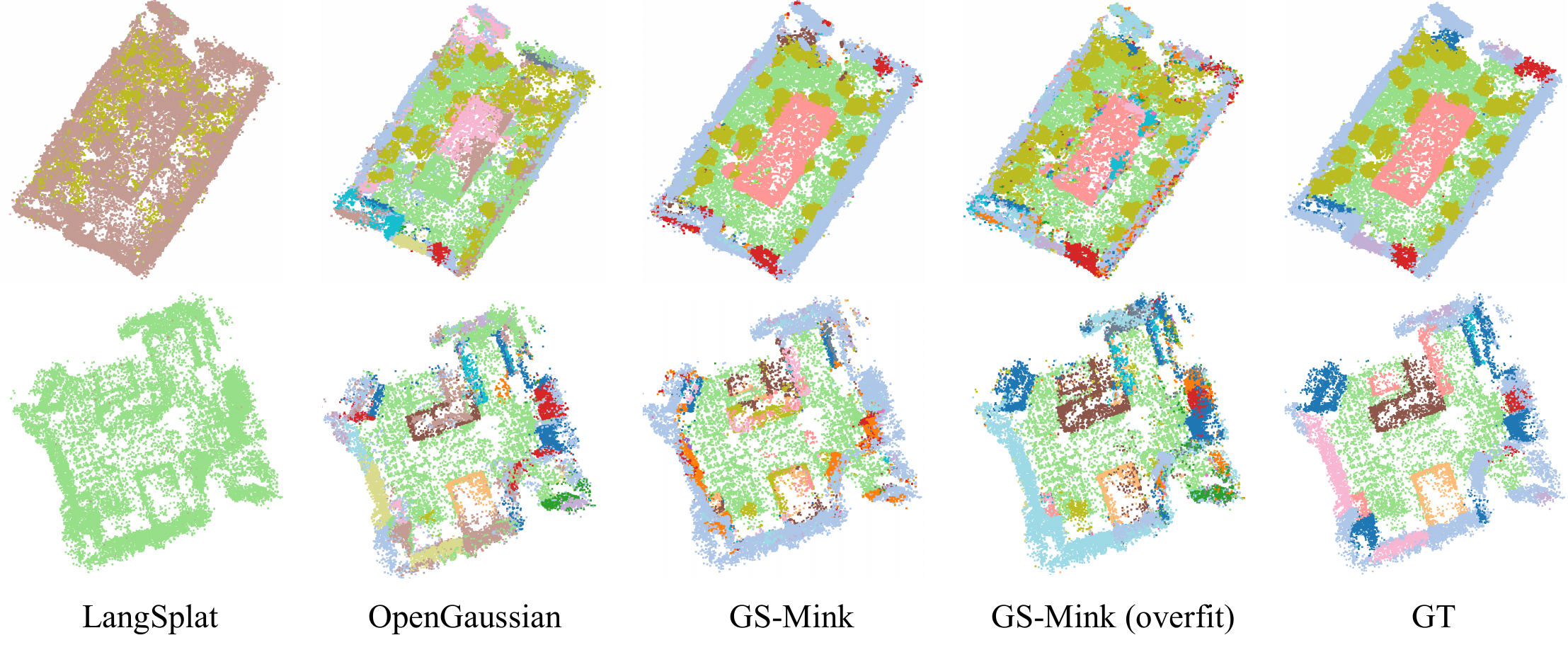}
    \caption{Qualitative results on the open-vocabulary 3D semantic segmentation task.}
    \label{fig:openvocab.qual}
\end{figure*}

%% file: table/openvocab.lerf_ovs.tex
\begin{table}[t] 
    \caption{Open-vocabulary semantic segmentation performances in the LeRF-OVS dataset.}
    \centering
    \resizebox{0.7\textwidth}{!}{
    \begin{tabular}{c|c|cc}
        \toprule
        Method & Feature dim. & \mIoU & \mAcc \\
        \midrule
        LangSplat~\citep{langsplat} & 3 & 9.7 & 12.4  \\ 
        LEGaussians~\citep{legaussian} & 8 & 16.2 & 23.8 \\
        OpenGaussian~\citep{opengaussian} & 6 & 38.4 & 51.4 \\
        SuperGSeg~\citep{liang2024supergseg} & 64 & 35.9 & 52.0 \\
        \midrule
        GS-Mink (ours) & 6 & 38.6 & 52.3  \\
        GS-Mink (ours) & 512 & \revision{\textbf{45.8}} & \revision{\textbf{56.9}} \\
        \bottomrule
    \end{tabular}
    }
    \label{table:lerf_dataset}
\end{table}

%% file: figure/6_lerf_ovs_qual/lerf_ovs_qual.tex
\begin{figure}[t]
    \centering
    \includegraphics[width=1.0\linewidth]{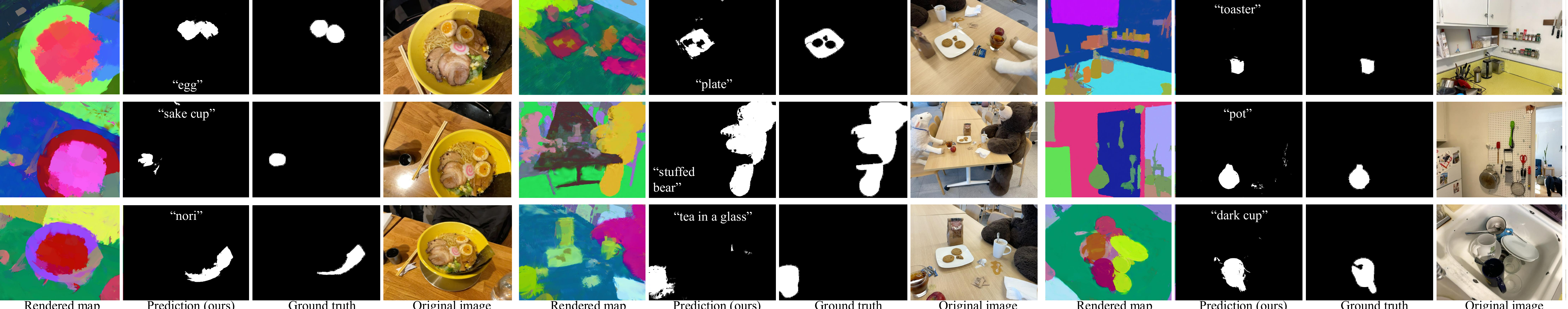}
    \caption{Our qualitative results in the LeRF-OVS dataset.}
    \label{rebutal.fig.rendered_featmap}
\end{figure}

%% file: figure/4_fps_miou/fps_miou.tex
\begin{figure}[!t]
    \centering
    \begin{minipage}{0.30\linewidth}
        \centering
        \includegraphics[width=1.0\linewidth]{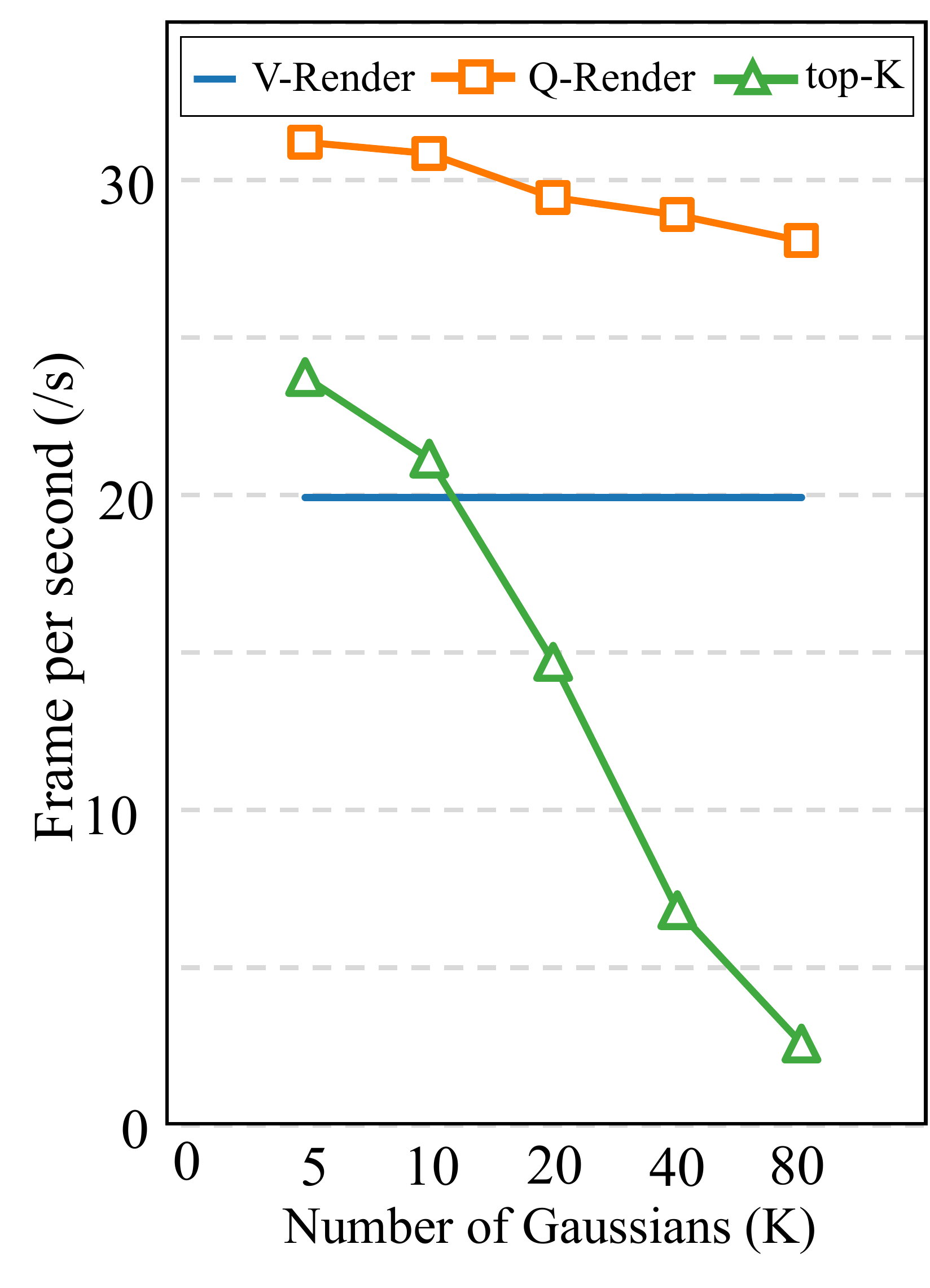}  
        \vspace{-2mm}
    \end{minipage}%
    \hfill
    \begin{minipage}{0.65\linewidth}
        \resizebox{1.0\linewidth}{!}{
        \begin{tabular}{c|c|c|c}
            \toprule
            
            3D neural network & Feature renderer & $K$ & 19 classes \mIoU \\
            \midrule
            
            \multicolumn{1}{c|}{\multirow{13}{*}{MinkUNet}} & V-Render~\citep{3dgs} & - & 49.02\\
            \cmidrule{2-4}
            

            & \multicolumn{1}{c|}{\multirow{5}{*}{top-$K$~\citep{dr_splat}}} & 5 & 37.84 \\
            & & 10 & 40.22 \\
            & & 20 & 43.59 \\
            & & 40 & 45.70 \\
            & & 50 & 44.93 \\
            \cmidrule{2-4}
            

            & \multicolumn{1}{c|}{\multirow{5}{*}{\samplernameshort (ours)}} & 5 & 49.98 \\
            & & 10 & 50.75 \\
            & & 20 & 50.65 \\
            & & 40 & \textbf{50.85} \\
            & & 50 & 50.28 \\
            \bottomrule
        \end{tabular}}
    \end{minipage}
    \vspace{-6mm}
    \caption{
        Comparison of (a) FPS and (b) mIoU by varying the number of Gaussians participating on rendering. Our \samplernameshort allows real-time rendering while preserving the mIoU performance. For the fair speed comparison, while using our implementation with the same rasterized Gaussians and 512-D Gaussian features, we only change the sampling strategies among the rasterized Gaussians: V-Render~\citep{nerf,3dgs} uses all intersecting Gaussians, top-$K$~\citep{dr_splat} sorts and selects $K$ Gaussians, and ours collect $K$ Quantile Gaussians.
    }
    \label{fig:render speed}
\end{figure}

%% file: table/ablation.grid_size.tex
\begin{table*}[!t]
    \caption{
        Ablation study for the grid size. Note that the scene scale is aligned metric-unit as described in~\secref{sec:Up-to-scale problem} of the appendix. So the grid size is determined as below.
    }
    \centering
    \resizebox{1.0\linewidth}{!}{
    \begin{tabular}{c|c|c|cc|cc|cc}
    \toprule
    \multicolumn{1}{c|}{\multirow{2}{*}{3D neural network}} & \multicolumn{1}{c|}{\multirow{2}{*}{Feature renderer}} & Grid Size & \multicolumn{2}{c|}{19 classes} & \multicolumn{2}{c|}{15 classes} & \multicolumn{2}{c}{10 classes}  \\
     & & (cm) & \mIoU & \mAcc & \mIoU & \mAcc & \mIoU & \mAcc \\
    \midrule

    \multirow{6}{*}{GS-Mink} & \multirow{9}{*}{\begin{tabular}[c]{@{}c@{}}Q-Render\\ (K=40)\end{tabular}} & 10.0 & 47.07 & 58.46 & 49.38 & 62.85 & 59.79 & 75.33 \\
    & & 5.0 & \textbf{50.39} & \textbf{62.14} & \textbf{53.14} & \textbf{66.39} & \textbf{63.37} & 78.30 \\
    & & 2.0 & 50.28 & 61.58 & 53.00 & 65.98 & 63.29 & \textbf{78.34} \\
    & & 1.0 & 45.00 & 55.41 & 48.27 & 60.99 & 60.29 & 75.45 \\
    & & 0.5 & 41.12 & 49.72 & 44.40 & 55.32 & 57.05 & 71.46 \\
    & & 0.25 & 34.36 & 42.42 & 37.34 & 47.47 & 49.03 & 61.78 \\ 
    \cmidrule{1-1}
    \cmidrule{3-9}
    
    \multirow{3}{*}{GS-PTv3} & & 10.0 & 43.71 & 55.96 & 49.94 & 63.09 & 59.60 & 74.79 \\
    & & 5.0 & 48.64 & 59.42 & 51.35 & 64.83 & 62.57 & 78.18 \\
    & & 2.0 & 48.99 & 60.36 & 52.39 & 66.05 & 62.57 & 77.70 \\
    \bottomrule

    \end{tabular}}
    \label{tab.grid_size}
    \vspace{-2mm}
\end{table*}

%% file: table/rebuttal.inference_time.tex
\begin{table}[!t]
    \caption{
        Rendering speed on ScanNet scene0006\_00 (frame 0). Note that 512\textsuperscript{\textdagger} is implemented by for-loop iterations, leveraging the original baseline code to render a 512 dimension feature map.
    }
    \centering
    \resizebox{0.8\linewidth}{!}{
    \begin{tabular}{c|ccc|ccc|ccc}
    \toprule
    
    & \multicolumn{3}{c}{LangSplat} & \multicolumn{3}{|c}{OpenGaussian} & \multicolumn{3}{|c}{GS-Mink (ours)} \\    
    \midrule
    
    Feature dim. & 3 & 6 & 512\textsuperscript{\textdagger} & 3 & 6 & 512\textsuperscript{\textdagger} & 3 & 6 & 512 \\
    \midrule
    
    FPS ($\uparrow$) & 112.12 & - & 0.65 & - & 71.13 & 0.83 & 172.52 & \textbf{80.98} & \textbf{28.42} (${\times}43.7{\uparrow}$) \\
    \bottomrule

    \end{tabular}}
    \label{rebuttal.table.inference_time}
\end{table}

%% file: figure/7_info_loss/info_loss.tex
\begin{figure}[!t]
    \centering
    \includegraphics[width=\linewidth]{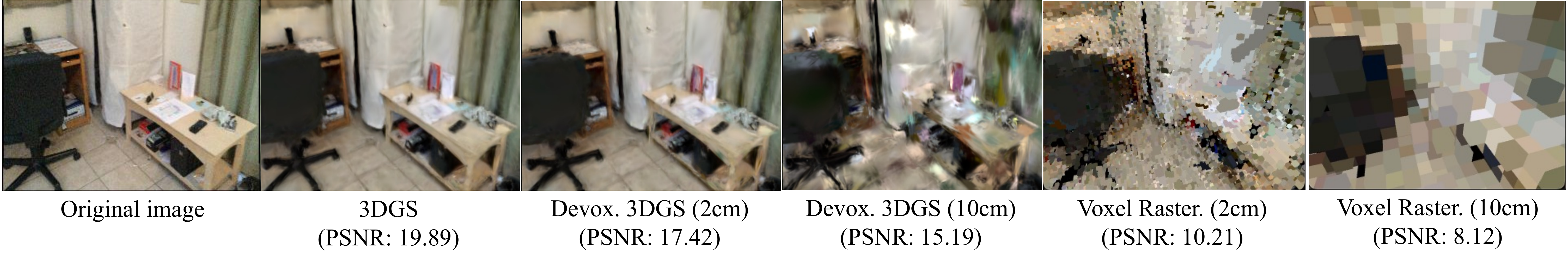}
    \vspace{-6mm}
    \caption{
        Information loss after voxelization. It shows that `\textit{a rendered image from de-voxelized Gaussians}' achieved higher fidelity compared to `\textit{a rendered image directly from sparse voxels}' (w/o de-voxelization).
    }
    \label{rebuttal.fig.info_loss}
\end{figure}

%% file: section/6_conclusion.tex
\section{Conclusion}

\revision{
The core contribution of this work is Quantile Rendering (Q-Render), a transmittance-aware strategy that resolves the computational bottleneck of embedding high-dimensional features in 3D Gaussian Splatting. Unlike conventional rendering that densely accumulates all intersections, Q-Render adaptively samples only the influential `quantile' Gaussians that dominate a ray's transmittance profile. This sparse sampling drastically reduces computational overhead, enabling efficient rendering of full 512-dimensional feature maps with up to a 43.7x speedup. 
Consequently, Q-Render achieves state-of-the-art performance on open-vocabulary segmentation benchmarks, establishing it as a scalable bridge between 2D foundation models and 3D representations. 
}

\revision{
\noindent \textbf{Limitations and Future Work.} 
Although Q-Render has shown efficient approximation of volumetric rendering, there still exists several limitations and failure cases.
}

\revision{
\noindent \textbf{Limitation 1: Dynamic $K$ Selection.}
Theoretically, the optimal number of samples $K$ should vary depending on the distribution of transmittance along a ray. 
However, in this work, we use a fixed $K$ across all experiments for simplicity. 
To investigate the sensitivity of our model to $K$, we conducted an analysis where we trained the model with $K=40$ and performed inference with varying $K \in \{5, 10, 20, 40, 50\}$. 
As shown in Table~\ref{tab:ablation_k_limit}, the performance significantly drops when the inference $K$ differs from the training configuration. 
This underscores the necessity of an adaptive $K$ selection strategy.
Although we explored two adaptive strategies in Appendix~\ref{subsec:appendix_adaptive}, they incur high computational costs. 
Thus, we leave the development of efficient, adaptive sampling strategies for future work.
}

\begin{table}[!t]
    \centering
    \caption{Performance change on the number of $K$. We use a model trained with $K=40$ and use different numbers of $K$ during inference. }
    \resizebox{0.52\textwidth}{!}{
    \begin{tabular}{lccccc}
        \toprule
        $K$ & 5 & 10 & 20 & 40 & 50 \\
        \midrule
        \mIoU & 39.16 & 42.18 & 44.94 & 45.81 & 45.71 \\
        \mAcc & 48.43 & 53.94 & 56.14 & 56.87 & 56.94 \\
        \bottomrule
    \end{tabular}
    }
    \label{tab:ablation_k_limit}
\end{table}

\begin{table*}[!t]
    \centering
    \caption{Comparison of GS-Net performance across different input 3D-GS models. GS-Net v2 utilizes 3D Gaussians pre-trained with an additional depth loss as input.}
    \label{tab:ablation_depth}
    \resizebox{1.00\textwidth}{!}{
    \begin{tabular}{l|c|cc|cc|cc}
        \toprule
        \multirow{2}{*}{Method} & \multirow{2}{*}{Training 3D-GS} & \multicolumn{2}{c|}{19 classes} & \multicolumn{2}{c|}{15 classes} & \multicolumn{2}{c}{10 classes} \\
        \cmidrule(lr){3-4} \cmidrule(lr){5-6} \cmidrule(lr){7-8}
         & & \mIoU & \mAcc & \mIoU & \mAcc & \mIoU & \mAcc \\
        \midrule
        GS-Net v1 & Rendering & 28.42 & 38.85 & 31.02 & 44.02 & 42.58 & 57.92 \\
        \textbf{GS-Net v2 (Ours)} & Rendering + Depth & \textbf{50.75} & \textbf{62.00} & \textbf{53.54} & \textbf{66.39} & \textbf{64.94} & \textbf{79.34} \\
        \bottomrule
    \end{tabular}
    }
\end{table*}

\revision{
\noindent \textbf{Limitation 2: Dependence on 3D-GS.}
Our current framework assumes that input 3D Gaussians are obtained through per-scene optimization, which inherently limits practical scalability. 
However, emerging generalizable 3D-GS approaches that eliminate the need for per-scene optimization, such as DepthSplat~\citep{xu2025depthsplat}, WorldMirror~\citep{Liu2025WorldMirror}, and DepthAnything3~\citep{DepthAnythingV3} offer a promising path to resolve this issue. 
Furthermore, we observe that the quality of the input 3D Gaussians significantly impacts downstream performance. 
As shown in Table~\ref{tab:ablation_depth}, GS-Net v1, which utilizes the original 3D-GS optimized without depth supervision, yields suboptimal results.
Conversely, GS-Net v2 takes as input a more recent 3D-GS implementation trained with additional depth loss. 
This improvement leads to substantial performance gains, highlighting the critical importance of geometric accuracy in the input representation.
We belive advancements in 3D-GS will also involve the improvement of GS-Net.
}

\revision{
\noindent \textbf{Limitation 3: Dependence on 3D Network Architecture.} 
As reported in Appendix~\ref{appendix:network_arch}, we observe that performance highly depends on the choice of backbone network. 
In particular, MinkowskiNet~\citep{mink} and PTv3~\citep{ptv3} exhibit strong sensitivity to the voxel grid resolution, as shown in Table~\ref{tab.grid_size}. 
This suggests that developing a more efficient and Gaussian-aware architecture could further improve the performance of our GS-Net framework. 
Potential directions include introducing Gaussian-friendly operators, reducing or eliminating voxelization, and designing modules that are robust to noise in Gaussian parameters.
}

%% file: appendix/a_up_to_scale.tex
\section{Resolving up-to-scale with monocular depth}
\label{sec:Up-to-scale problem}

The scale of a 3D scene is a critical factor in 3D scene understanding. However, when camera poses are estimated solely from multi-view images, recovering the absolute scene scale—referred to as the up-to-scale problem—becomes challenging. For instance, COLMAP~\cite{colmap_0,colmap_1}, a widely used tool for extracting extrinsic and intrinsic camera parameters in neural rendering pipelines, does not inherently address this issue. Since our method relies on optimized Gaussian parameters~$\Theta$, the up-to-scale problem naturally arises when handling diverse 3D scenes captured purely from images.

To overcome this issue, we employ DepthAnythingV2~\cite{depthanythingv2}, an off-the-shelf monocular depth estimation method trained to produce metric-scale depth maps from single images. Similar to the depth alignment process in ~\cite{hierarchical3dgs, depthreg_3dgs}, we fix the original Gaussian parameters~$\Theta$ and optimize the global scene scale~$a \in \Real$ to approximate the metric scale:
\begin{equation}
    \mathop{\arg\min}_{a} \sum_{i=1}^{N_\mathcal{I}} | \text{invD}^\text{mono}_i - a \cdot \text{invD}^\text{rendered}_i |_1,
\end{equation}
where $N_\mathcal{I}$ is the number of known images, $\text{invD}^\text{mono}_i$ is $i$-th inverse depth map from DepthAnythingV2~\cite{depthanythingv2}, and $\text{invD}^\text{rendered}_i$ is the corresponding rendered inverse depth map using 3DGS.
After obtaining the scene scale~$a$, we update the Gaussian parameter~$\Theta$ as $\tilde{\Theta} = \set{\tilde{\theta}_i}_{i=1}^{N} = \set{\frac{1}{a}\cdot\bmu_i, \frac{1}{a}\cdot\scale_i, \rot_i, \alpha_i, \sph_i}_{i=1}^N$. Note that only the mean and the scale attributes of each Gaussian are changed. We provide an example result in~\Figref{fig:scene_scale}.

Unlike the original 3DGS~\cite{3dgs}, which recently introduced a scale-alignment method, we avoid pruning initial COLMAP points that could affect the distribution of our optimized Gaussian parameters~$\Theta$. 
This ensures a fair comparison with recent studies such as Gaga~\cite{gaga}. 
Furthermore, we use a metric depth estimator, \textit{`depth-anything-v2-metric-hypersim-vitl.pth`} in the official repository, whereas 3DGS employs an inverse depth estimator, \textit{`depth-anything-v2-vitl.pth'}. 
Instead of normalizing the estimated depth maps, we directly take their reciprocal to obtain inverse depth maps~$D^\text{mono}$.

%% file: figure/a_scene_scale/scene_scale.tex
\begin{figure*}[!htb]
    \centering
    \includegraphics[width=0.7\linewidth]{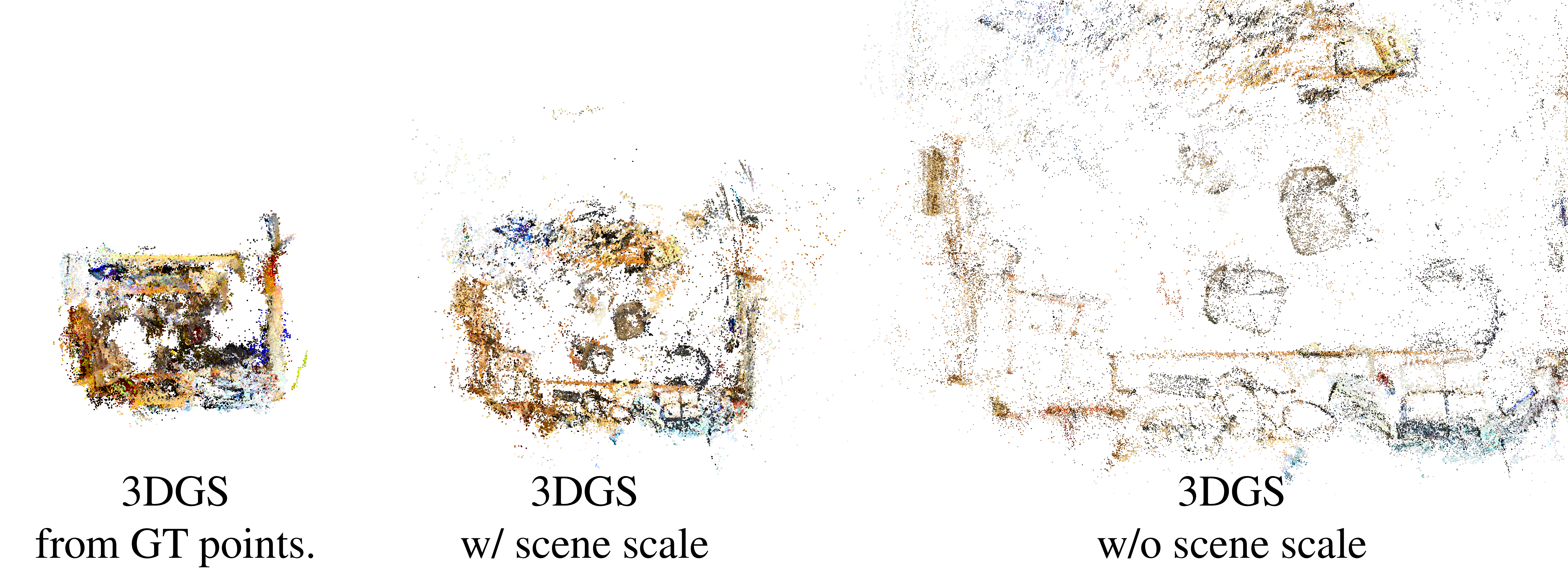}
    \caption{Visualization of 3D Gaussians with different conditions: (left) 3D Gaussians that are trained from ground truth pointcloud, (middle) 3D Gaussians that are trained from COLMAP points, then are applied with scene scale~$a$, (right) 3D Gaussians that are trained from COLMAP points without considering scene scale~$a$.}
    \label{fig:scene_scale}
\end{figure*}

%% file: appendix/b_data_preprocessing.tex
\section{Data preprocessing}
\label{sec:Data preprocessing}

\subsection{3D Gaussian Splatting optimization}
\label{subsec:Optimize 3D Gaussian Splattings}
We use the original implementation with the proposed hyperparameters, such as training iterations and upsampling frequency, for pre-optimizing 3D-GS.
For the initial points, we follow the initialization strategies used by OpenGaussian~\cite{opengaussian} and Dr.Splat~\cite{dr_splat}, utilizing the point clouds provided by the original ScanNetv2 dataset instead of using COLMAP~\cite{colmap_0, colmap_1}.
Note that we maintain the densification and pruning processes of 3D-GS to preserve its capacity.

\subsection{Pseudo ground truth acquisition}
\label{subsec:Pseudo ground truth acquisition}
Since we enable the densification and pruning processes of 3D-GS, the number and coordinates of each Gaussian differ from the original point clouds. 
To evaluate the predicted Gaussians, we need to create Gaussian-specific labels derived from the original point labels.
As shown in \Figref{fig:gt_conversion}, for each Gaussian, we first explore the $K$ nearest neighbor points and assign the label based on the most frequent label among these neighbors. 
However, we observed that due to the large number of Gaussians, many Gaussians contribute minimally to the scene (i.e., their opacity is below 0.1), and we filtered out these less significant Gaussians.
Despite this, we found that some isolated Gaussians (i.e., "floaters") still received labels, resulting in noisy ground truth (GT) labels. 
To address this, we ignored labels if all the nearest points had a Mahalanobis distance above 0.1, effectively removing such noisy assignments. 
As a result, we were able to obtain clearer and more consistent Gaussian labels.

%% file: figure/b_gt_label/gt_label.tex
\begin{figure*}[t!]
    \centering
    \includegraphics[width=\linewidth]{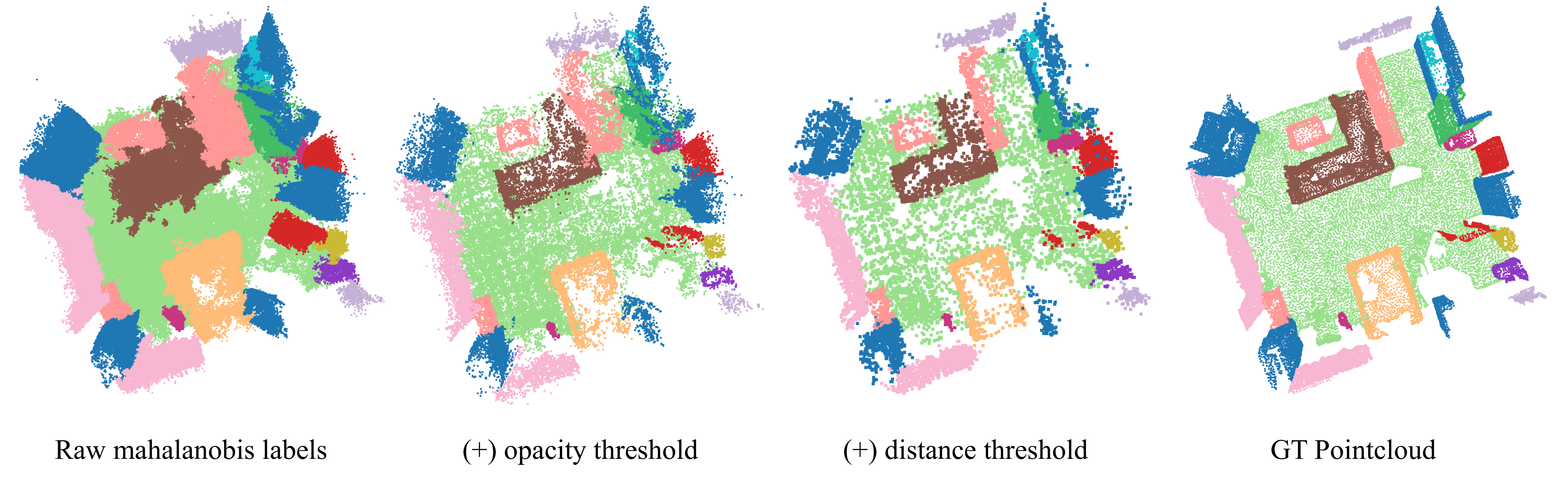}
    \caption{Generated Gaussian Splatting labels from point cloud labels. Applying both opacity and distance thresholding allows much clearer label generation for Gaussian Splatting.}
    \label{fig:gt_conversion}
\end{figure*}

%% file: appendix/c_theory.tex
\section{Theoretical Analysis of Quantile Rendering}
\label{section:theortical_derivation}

\revision{
Based on the definitions provided in Section 4.2 and Algorithm 1 of the manuscript, we provide the theoretical analysis of Quantile Rendering (Q-Render) as an approximation of the volume rendering in 3D-GS~\citep{3dgs} using Riemann sums.
}


\subsection{Preliminaries: Riemann Sums and the Right Rule}
\revision{
Let $f: [a, b] \to \mathbb{R}$ be a bounded function defined on a closed interval $[a, b]$. We define a partition $P$ of the interval into $N$ sub-intervals:
\[
a = u_0 < u_1 < u_2 < \dots < u_N = b
\]
The width of the $k$-th sub-interval $[u_{k-1}, u_k]$ is denoted by $\Delta u_k = u_k - u_{k-1}$. A \textbf{Riemann Sum} approximates the definite integral of $f$ over $[a, b]$ by summing the areas of rectangles:
\begin{equation}
    S = \sum_{k=1}^{N} f(u_k^*) \Delta u_k
\end{equation}
where $u_k^*$ is a sample point chosen from the sub-interval $[u_{k-1}, u_k]$.
}

\revision{
The \textbf{Right Riemann Sum} specifically chooses the right endpoint of each sub-interval as the sample point:
\begin{equation}
    u_k^* = u_k.
\end{equation}
Thus, the approximation becomes:
\begin{equation}
    S_{right} = \sum_{k=1}^{N} f(u_k) \Delta u_k
\end{equation}
}

\revision{
When the partition is uniform, $\Delta x = (b-a)/N$, due to the `Right endpoint approximation theorem', the Right Rule converges linearly:
\[
\left|\int_a^b f(x)\,dx - S_{\mathrm{right}}\right|
\le \frac{M(b-a)^2}{2N},
\]
where $M$ is the maximum value of the absolute value of $\frac{\partial f(x)}{\partial x}$ over the interval. 
}


\subsection{Volume Rendering as an Integral}

\revision{
We observe that the discrete volume rendering process is fundamentally equivalent to a Riemann sum approximation of a continuous path integral. Theoretically, volume rendering assumes that the accumulated transmittance $T$ monotonically decreases from $1$ to $0$ along the ray, creating a bounded integration domain. Different neural rendering approaches can be characterized by how they partition this domain:
\begin{itemize}
    \item \textbf{NeRF~\citep{nerf}:} The integral is approximated in the \textit{spatial domain} ($t$). Query points are sampled uniformly or hierarchically (coarse-to-fine) along the ray. The Riemann partition is defined by the distance between adjacent samples ($\delta_i = t_{i+1} - t_i$), where the opacity $\alpha_i$ is derived from the learned volumetric density $\sigma$.
    \item \textbf{3D-GS~\citep{3dgs}:} The integral is evaluated using a set of discrete primitives sorted by depth. Here, the rasterized Gaussians act as the representative samples for intervals along the ray. The rendering equation accumulates the alpha-blended contribution of each intersecting Gaussian, effectively performing a Riemann summation where the step size and weight are determined by the Gaussian's covariance and opacity.
    \item \textbf{Quantile Rendering (Ours):} Unlike the spatial sampling in NeRF or the dense accumulation in 3D-GS, our method shifts the Riemann partition to the \textit{transmittance domain} ($T$). By sampling intervals of equal probability in transmittance space, we ensure that the summation is performed only over the most significant contributors, offering a more efficient approximation of the underlying integral.
\end{itemize}
}


\subsection{Volume Rendering as a Continuous Integral}

\revision{
Consider a camera ray parameterized by distance $t \in [0,\infty)$.  
Let $\sigma(t)$ denote the volume density and $c(t)$ the feature (or color) emitted at position $t$. 
}

\revision{
The transmittance is defined as
\[
T(t) = \exp\!\Big(-\int_0^t \sigma(s)\,ds\Big).
\]
Its derivative satisfies
\[
\frac{dT}{dt} = -\sigma(t) T(t)
\quad\Longrightarrow\quad
-\frac{dT}{dt}\,dt = \sigma(t)T(t)\,dt.
\]
}

\revision{
The continuous volume rendering integral is
\begin{equation} \label{eq:vol-t-integral}
    C_{\mathrm{vol}} = \int_{0}^{\infty} c(t)\,\sigma(t)T(t)\,dt.
\end{equation}
}

\revision{
The Quantile render operates by analyzing the change in Transmittance $T$ rather than spatial distance~$t$. We can reformulate the volume rendering integral Eq.~\ref{eq:vol-t-integral} by performing a change of variables from spatial distance $t$ to transmittance $u = T(t)$. Following \citep{nerf,3dgs}, $u$ and $t$ are noted as:
\begin{itemize}
    \item When $t=0$, $u=1$ (Ray starts with full transmittance starting from $1$).
    \item When $t \to \infty$, $u \to 0$ (Transmittance converges to $0$).
\end{itemize}
Since $T$ is strictly decreasing from $1$ to $0$, the substitution
\[
u = T(t), \qquad du = -\sigma(t)T(t)\,dt
\]
transforms Eq.~\ref{eq:vol-t-integral} into
\[
C_{\mathrm{vol}}
= \int_{u=1}^{0} c(u)\,(-du)
= \int_{0}^{1} c(u)\,du.
\]
}

\revision{
Thus, volume rendering is exactly the integral of a function $c(u)$ over the transmittance interval $[0,1]$:
\begin{equation}
C_{\mathrm{vol}} = \int_0^1 c(u)\,du.
\label{eq:vol-u-integral}
\end{equation}
}


\subsection{Quantile Rendering as a Right Riemann Sum}
\revision{
As described in Section 4.2 of the manuscript, the Quantile Rendering algorithm partitions the Transmittance $T \in [0, 1]$ into $K+1$ evenly distributed segments. The algorithm selects a specific `Quantile Gaussian' whenever the accumulated transmittance drops. 
}

\revision{
In detail, the Quantile Rendering partitions the transmittance domain $[0,1]$ into
\[
0 = u_{K+1} < u_{K} < \cdots < u_1  < u_0 = 1.
\]
For each interval $[u_{k-1}, u_k]$, Q-render selects the Gaussian whose 
transmittance crossing corresponds to the right endpoint $u_k$.  
Thus the algorithm evaluates $c(u)$ at $u_k$. Therefore, Q-render implements exactly the Right Riemann Sum for Eq.~\ref{eq:vol-u-integral}:
\begin{equation} \label{eq:q-right}
    C_{Q}^{\mathrm{right}} = \sum_{k=1}^{K+1} c(u_k)\,\Delta u.
\end{equation}
This replaces the dense per-Gaussian accumulation with a sparse, quantile-driven sampling strategy.
}


\subsection{Approximation Error of Quantile Rendering}
\revision{
Following~\citep{nerf,3dgs}, the integrand $c(u)$ is differentiable and satisfies
\[
|c'(u)| \le M \quad \forall u\in[0,1].
\]
Applying the classical Right Rule error bound with $a=0$, $b=1$, and $N=K+1$ gives:
\begin{equation} \label{eq:right-bound}
\big| C_{\mathrm{vol}} - C_Q^{\mathrm{right}} \big|
\le \frac{M}{2(K+1)}
\le \frac{M}{2K}.
\end{equation}
Thus the quantile approximation converges at rate $O(1/K)$.
}

\subsection{Convergence of Q-render with Transmittance Normalization}

\revision{
Q-render includes a final normalization step to correct for residual transmittance:
\begin{equation}
    \widetilde C_Q = \frac{C_Q^{\mathrm{right}}}{1 - T_Q},    
\end{equation}
where $T_Q$ is the remaining transmittance after processing all $K$ quantile Gaussians.
}

\revision{
Since each quantile removes at least $\Delta u$, we have
\begin{equation}
    T_Q \le \Delta u = \frac{1}{K+1}.    
\end{equation}
Hence the normalization factor satisfies
\begin{equation}
    \frac{1}{1 - T_Q}
\le \frac{1}{1 - \frac{1}{K+1}}
= \frac{K+1}{K}.    
\end{equation}
}

\revision{
Combining with \eqref{eq:right-bound} yields
\begin{equation}
    \big|C_{\mathrm{vol}} - \widetilde C_Q \big|
\le \frac{K+1}{K} \cdot \frac{M}{2(K+1)}
= \frac{M}{2K}.    
\end{equation}
}

\paragraph{Final Theorem.}
\revision{
Under the assumptions that $c$ is differentiable and $|c'(u)|\le M$~\citep{nerf,3dgs},  
Q-render converges to volume rendering with rate
\begin{equation}
\boxed{
\big|C_{\mathrm{vol}} - \widetilde C_Q \big|
\le \frac{M}{2K}
}
\end{equation}
and the approximation error vanishes linearly as $K\to\infty$.
}

%% file: appendix/d_implementation_details.tex
\section{Implementation Details}
For optimization, we use the default optimizers provided by MinkUNet (Adam) and PTv3 (AdamW). The learning rate is adjusted using PyTorch's ReduceLROnPlateau scheduler, reducing by a factor of 10 when a plateau is detected. Training is conducted with a batch size of 4 across 8 A100-80GB GPUs while computing rendering loss from 4 randomly chosen training viewpoints. Both MinkUNet and PTv3 follow their default configurations.


%% file: appendix/e_additional_exp.tex
\section{Additional Experiments}

\subsection{Adaptive Gaussian Sampling Strategy}
\label{subsec:appendix_adaptive}
\revision{
We additionally implement two adaptive variants of Q-Render, namely \texttt{Learned-K} and \texttt{Stratified-K}, to assess whether adaptive selection of $K$ can effectively mitigate the sensitivity to this hyperparameter.
}

\paragraph{Learned-K.}
\revision{
In this variant, we train the model with three candidate values for $K$, specifically $\{10, 20, 40\}$. 
Inspired by the mIoU-prediction head in SAM2, we introduce an additional \emph{similarity-prediction head} that estimates the expected similarity between the predicted features and the target features for each candidate $K$. 
This head takes the transmittance profile as input and outputs the predicted similarity scores.
Therefore, similar to SAM2, we use an additional loss that computes the difference between predicted similarity and the actual similarity. 
}

\begin{equation}
    \mathcal{L}_{sim}(\tilde{\mathbf{f}}^Q, \mathbf{f}_i^{\text{CLIP}}, \{T_n\}_{n=1}^N )  =  ||\operatorname{Sim-Head}_{\Theta}(\{T_n\}_{n=1}^N) - \operatorname{sim}(\tilde{\mathbf{f}}^Q, \mathbf{f}_i^{\text{CLIP}})||_2^2,
\end{equation}

\revision{
where $\{T_n\}_{n=1}^N$ are transmittance values along the ray, $\tilde{\mathbf{f}}^Q$ is a rendered feature via Q-Render, and $\mathbf{f}_i^{\text{CLIP}}$ is the CLIP feature used in training. 
}

\revision{
In addition, we apply the original loss for three $K$s during training to optimize features for all $K$s. 
During inference, the model evaluates these similarity scores and dynamically selects the value of $K$ that yields the highest expected similarity. 
This allows the model to adaptively determine $K$ per input without relying on a manually fixed value.
}

\paragraph{Stratified-sampling.}

\revision{
In the second variant, we compute the mean and standard deviation of the transmittance values along the depth dimension. 
Rather than uniformly partitioning the transmittance for sampling, we instead draw samples based on the \emph{z-score} under a Gaussian distribution parameterized by the estimated mean and standard deviation. 
This enables denser sampling at depths where objects are more likely to contribute, analogous to the stratified sampling strategy used in NeRF, and improves robustness under diverse transmittance distributions.
}
\revision{
Specifically, we uniformly partition the transmittance interval in the $z$-score space and then map the samples back to the original transmittance domain. Consequently, unlike the original Q-Render--which uniformly partitions the transmittance values--this variant adapts the sampling locations according to the underlying transmittance distribution, encouraging the model to draw more Gaussians from regions where they are more densely concentrated.
}

\paragraph{Results.}
\revision{
We compare these two adaptive variants against the original Q-Render in terms of segmentation performance and rendering FPS. Due to the large memory footprint of the \texttt{Learned-K} model, we use a voxel size of 0.5 for all experiments. For a fair comparison, we set $K=40$ for both the \texttt{stratified-K} variant and the original Q-Render.
As shown in Table~\ref{tab:adaptive_gaussian}, the \texttt{Learned-K} model slightly underperforms the other strategies, likely because the model must additionally learn the similarity-prediction head, which makes it harder to focus on improving feature quality. We also observe that the \texttt{stratified-K} variant performs on par with the original Q-Render.
However, the most critical observation is that both adaptive variants achieve nearly half the FPS of Q-Render. This slowdown arises because they require two passes along each ray: one to estimate the transmittance statistics and another to perform rendering. Despite their increased computational cost, we do not observe any clear performance improvement over Q-Render. Therefore, we adopt Q-Render as our rendering algorithm of choice, as it provides an efficient and effective approximation of V-Render.
}

\begin{table*}[!tb]
    \caption{
        Open vocabulary 3D semantic segmentation performances in the ScanNet dataset. 
    }
    \centering
    \resizebox{1.0\linewidth}{!}{
    \begin{tabular}{l|c|cc|cc|cc}
    \toprule
    \multicolumn{1}{c|}{\multirow{2}{*}{Method}} & \multicolumn{1}{c|}{\multirow{2}{*}{FPS}} & \multicolumn{2}{c|}{19 classes} & \multicolumn{2}{c|}{15 classes} & \multicolumn{2}{c}{10 classes} \\
    
     & & \mIoU & \mAcc & \mIoU & \mAcc & \mIoU & \mAcc \\
    \midrule
    Learned-K & 14.31 & 40.14 & 49.16 & 43.85 & 55.60 & 55.84 & 68.53  \\
    Stratified-K & 15.14 & \textbf{41.30} & 49.52 & \textbf{44.67} & \textbf{55.78} & 56.66 & 70.10  \\
    Q-Render (ours) & \textbf{32.17} & 41.12 & \textbf{49.72} & 44.40 & 55.32 & \textbf{57.05} & \textbf{71.46}   \\
    \bottomrule 
    \end{tabular}
    }
    \label{tab:adaptive_gaussian}
\end{table*}

\subsection{Open Vocabulary Segmentation on Outdoor Scenes}

\revision{
Since our evaluation benchmarks are all indoor datasets, we also demonstrate the effectiveness of our proposed method on outdoor benchmarks. 
However, to the best of our knowledge, there are currently no benchmarks that jointly cover multi-view and outdoor scenes for open-vocabulary segmentation (OVS). 
To enable comparison in outdoor settings, we manually annotated 3 scenes from MipNeRF360~\citep{mip_nerf_360} outdoor scenes. 
Specifically, we use SAM2 to obtain an initial mask corresponding to the queried object and then manually refine the segmentation by providing additional positive and negative point prompts. 
This process allows for reliable ground-truth masks for evaluating outdoor multi-view OVS performance across all baselines. 
}

\revision{
We compare our framework with the state-of-the-art method, Dr.Splat~\citep{dr_splat}, using the manually annotated MipNeRF360 benchmark.
As reported in Table~\ref{tab:outdoor}, Q-Render consistently outperforms Dr.Splat, corroborating our main findings and demonstrating robust generalization to outdoor scenarios. 
Qualitative comparisons are visualized in Figure~\ref{fig:mipnerf360}. As observed in the first column, our method achieves clear separation between the target object and its surroundings. 
However, we acknowledge that VRAM constraints necessitated the use of larger voxel sizes. 
We leave the development of more memory-efficient network architectures for 3D-GS to future work.
}

\begin{table}[!t]
\centering
\caption{Comparison between Dr.Splat and GS-Mink on MipNeRF360 benchmark. }
\begin{tabular}{l|cc|cc}
\toprule
\multirow{2}{*}{Scene} 
& \multicolumn{2}{c|}{Dr.Splat} 
& \multicolumn{2}{c}{GS-Mink} \\
 & mIoU & mAcc & mIoU & mAcc \\
\midrule
bicycle  & 0.2112 & 0.3012 & \textbf{0.2236} & \textbf{0.3165} \\
garden   & 0.5543 & 0.6114 & \textbf{0.6721} & \textbf{0.7813} \\
treehill & \textbf{0.2122} & \textbf{0.2713} & 0.2063 & 0.2585 \\
\midrule
Avg. & 0.3359 & 0.3946 & \textbf{0.3673} & \textbf{0.4521} \\
\bottomrule
\end{tabular}
\label{tab:outdoor}
\end{table}

\input{figure/e_additional_exp/mipnerf360}

\subsection{More network architecture}
\label{appendix:network_arch}

\revision{
To evaluate the generalizability of our approach across different backbones, we conducted additional experiments using two representative point-based architectures: PointNet++~\citep{pointnetpp} and PointNeXT~\citep{pointnext}. PointNet++ utilizes hierarchical feature learning via local neighborhood grouping to capture fine-grained geometry, while PointNeXT advances this design with residual connections and improved scaling strategies.
}
\revision{
However, as reported in Table~\ref{tab:model_arch}, both point-based models exhibit inferior performance compared to voxel-based baselines (MinkUNet and PTv3). We conjecture that this disparity stems from the handling of point density. Since 3D Gaussians are often densely clustered around object surfaces, the k-nearest neighbor (k-NN) search used in point-based networks tends to limit the metric receptive field, thereby hindering the aggregation of broader contextual information. In contrast, voxel-based networks are inherently more robust to such density variations, resulting in superior performance.
}

\begin{table*}[!tb]
    \caption{
        Comparison of various 3D network backbone to GS-Net.
    }
    \centering
    \resizebox{\linewidth}{!}{
    \begin{tabular}{l|c|cc|cc|cc}
    \toprule
    \multicolumn{1}{c|}{\multirow{2}{*}{3D arch.}} & \multirow{2}{*}{Rendering} & \multicolumn{2}{c|}{19 classes} & \multicolumn{2}{c|}{15 classes} & \multicolumn{2}{c}{10 classes} \\
    
     & & \mIoU & \mAcc & \mIoU & \mAcc & \mIoU & \mAcc \\
    \midrule
    PointNet++~\citep{pointnetpp} & \multirow{4}{*}{Q-render} & 39.42 & 50.62 & 42.88 & 56.66 & 52.21 & 69.14 \\  
    PointNeXT~\citep{pointnext} & & 37.89 & 45.80 & 41.15 & 52.09 & 51.57 & 63.83 \\  
    PTv3~\citep{ptv3} & & 48.99 & 60.36 & 52.39 & 66.05 & 64.95 & 79.34 \\
    MinkUNet~\citep{mink} & & 50.75 & 62.00 & 53.54 & 66.39 & 64.95 & 79.34 \\
    \bottomrule 
    \end{tabular}
    }
    \label{tab:model_arch}
\end{table*}

\subsection{Thorough Comparison with OpenGaussian}

\revision{
To better understand which components of our method contribute most to the performance improvement, we conduct an ablation by modifying each part of the pipeline individually and comparing the results against OpenGaussian. Specifically, we introduce two hybrid models: one that uses the GS-Net feature extractor while retaining the original rendering module of OpenGaussian, and another that keeps the OpenGaussian feature extractor but replaces the renderer with Q-Render. As shown in Table~\ref{tab:vq_render_compare}, incorporating Q-Render yields a consistent performance gain for both OpenGaussian and GS-Net. Moreover, replacing the OpenGaussian codebooks with GS-Net features leads to a substantial improvement. We conjecture that preserving high-dimensional features is a key factor, as OpenGaussian encodes only 6-dimensional attributes, whereas GS-Net leverages 512-dimensional CLIP features that retain richer semantic information.
}

\begin{table}[t]
\centering
\caption{Comparison of V-Render and Q-Render under different feature extractors.}
\resizebox{\textwidth}{!}{
\begin{tabular}{l|l|cccccc}
\toprule
\multirow{2}{*}{Feature Extractor} & \multirow{2}{*}{Rendering} & \multicolumn{2}{c}{19 classes} & \multicolumn{2}{c}{15 classes} & \multicolumn{2}{c}{10 classes} \\
& & \mIoU & \mAcc & \mIoU & \mAcc & \mIoU & \mAcc \\
\midrule
\multirow{2}{*}{OpenGaussian}
& V-Render & 22.60  & 34.41 & 24.21 & 37.58 & 34.74 & 51.58 \\
& Q-Render & 23.10  & 35.55 & 26.18 & 38.12 & 36.13 & 56.12 \\
\midrule
\multirow{2}{*}{GS-Net (Mink)}
& V-Render & 49.02 & 58.75 & 50.41 & 63.65 & 61.04 & 76.21 \\  
& Q-Render & 50.75 & 62.00    & 53.54 & 66.39 & 64.94 & 79.34 \\
\bottomrule
\end{tabular}
}
\label{tab:vq_render_compare}
\end{table}

\subsection{Sensitivity under Gaussian Perturbation}

\revision{
We also investigate the sensitivity of GS-Net to perturbations applied to the input 3D-GS representation. Specifically, on top of the predicted opacity, we inject Gaussian noise with varying scales into the Gaussian opacity values and evaluate the resulting segmentation performance. 
To avoid extreme cases where the opacity becomes negative values, we apply the noise before the sigmoid activation is applied.
As shown in Table~\ref{tab:qrender_noise}, GS-Net remains highly robust under mild noise levels: performance degradation at noise scales of 0.25 and 0.5 is negligible, demonstrating that Q-Render retains sufficient stability even when the opacity field is moderately corrupted. 
However, as the noise scale increases beyond 1.0, we observe a marginal drop in both mIoU and mAcc, and the performance eventually collapses at extreme noise levels such as 4.0. 
These results indicate that while Q-Render is resilient to realistic perturbations in the opacity field, strong distortions that heavily corrupt the underlying geometry inevitably degrade performance. 
Overall, this experiment suggests that our pipeline maintains robust performance under practical levels of noise in 3D-GS inputs.
}

\begin{table}[t]
\centering
\caption{Robustness of Q-Render to Gaussian noise added to opacity.}
\begin{tabular}{c|cc}
\toprule
\multirow{2}{*}{Noise Scale} & \multicolumn{2}{c}{19 classes} \\
& \mIoU & \mAcc \\
\midrule
0    & 50.75 & 62.00 \\
0.25 & 50.13 & 60.94 \\
0.5  & 50.18 & 60.88 \\
1    & 47.13 & 57.10 \\
2    & 38.13 & 44.11 \\
4    & 16.12 & 32.19 \\
\bottomrule
\end{tabular}
\label{tab:qrender_noise}
\end{table}

\subsection{Rendering RGB}

\revision{
Since our Q-Render is not restricted to CLIP features, it could also be applied to RGB image rendering. 
Using this setup, we quantify how much information is lost when replacing V-Render with Q-Render. To isolate the behavior of the renderer itself, we directly apply Q-Render to pre-trained 3D-GS models without any fine-tuning. 
As shown in Fig.~\ref{fig:qrgb}, Q-Render exhibits only a very slight drop in PSNR compared to V-Render, indicating that it provides an effective approximation of the original renderer even in the RGB domain. 
This suggests that Q-Render is broadly applicable beyond our semantic feature rendering setup. In particular, we believe Q-Render can be seamlessly integrated into various 3D-GS variants—including dynamic or time-varying Gaussians—since these models also assume fixed Gaussian positions at each rendering step.
}

\input{figure/11_qrender_rgb/11_qrender_rgb}

\subsection{Memory Footprint Comparison}

\revision{
We also compare the VRAM requirements of each baseline on \texttt{scene0000\_00} from ScanNet (100 frames). As summarized in Table~\ref{tab:memory_compare}, LangSplat and OpenGaussian require little memory due to their low-dimensional feature representations (3D and 6D). However, this compression inevitably introduces information loss, which correlates with their reduced segmentation performance. 
}

\revision{
Dr.Splat relies on 512D features and stores per-view Gaussian visibility masks, which causes memory usage to grow proportionally with the number of frames and leads to more than 61,GB of peak consumption. Q-Render also uses 512D features, and its cache-free per-ray accumulation avoids this overhead, reducing the peak memory to 27.18,GB at $K{=}40$ while preserving high-dimensional semantic capacity.
}

\begin{table}[t]
\centering
\caption{Peak inference memory on \texttt{scene0000\_00} (ScanNet, 100 frames).}
\begin{tabular}{l|c|c}
\toprule
Method & Feature Dim. & Peak Memory (GB) \\
\midrule
LangSplat & 3   & 7.18  \\
OpenGaussian & 6 & 16.13 \\
Dr.Splat & 512 & 61.13 \\
Q-Render ($K{=}40$) & 512 & 27.18 \\
\bottomrule
\end{tabular}
\label{tab:memory_compare}
\end{table}

%% file: figure/e_additional_exp/mipnerf360.tex
\begin{figure*}[t!]
    \centering
    \includegraphics[width=\linewidth]{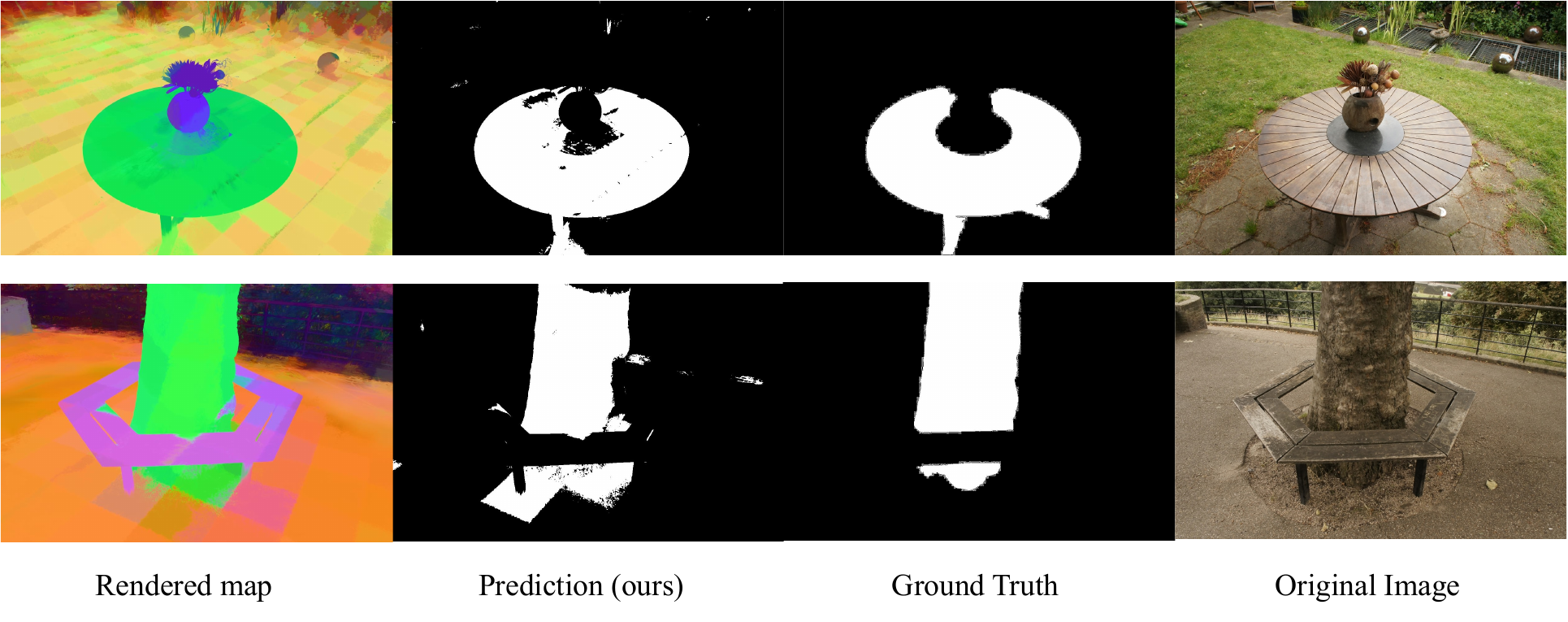}
    \caption{Our qualitative results in the MipNeRF360 dataset.}
    \label{fig:mipnerf360}
\end{figure*}

%% file: figure/11_qrender_rgb/11_qrender_rgb.tex
\begin{figure*}[t!]
    \centering
    \includegraphics[width=1.0\linewidth]{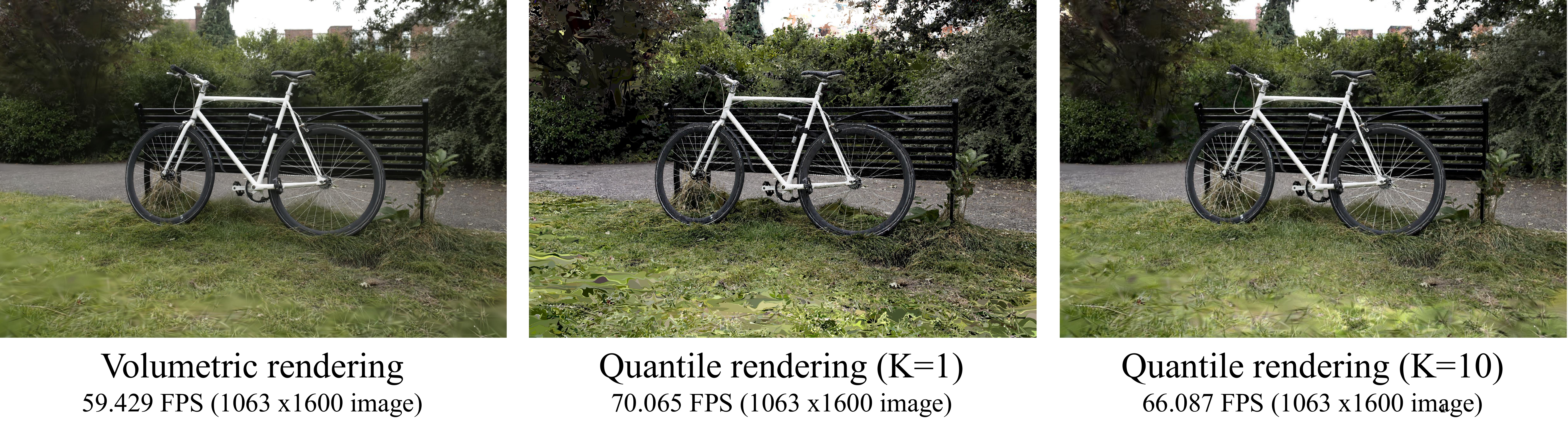}
    \caption{
    Comparison between V-Render and Q-Render on RGB reconstruction using a pre-trained 3D-GS model without fine-tuning. Q-Render shows only minor PSNR degradation, demonstrating that it serves as an efficient and accurate approximation of V-Render in the RGB domain.
    }
    \label{fig:qrgb}
\end{figure*}

%% file: appendix/f_pseudocode.tex
\section{Pseudo-code for Voxelization, Voxel Feature Composition, and De-voxelization}

\revision{
To ensure reproducibility, we provide detailed pseudo-code of our implementation. Listing~\ref{code:voxelization} outlines the voxelization pipeline that groups Gaussians into voxels for efficient processing. 
Listing~\ref{code:compose_voxel_features} details the organization of voxel positions and features as inputs for the 3D network. 
Finally, Listing~\ref{code:de_voxelization} describes the de-voxelization process, where voxel features are redistributed to predict Gaussian-level features.
}

\input{algorithm/voxelization}
\input{algorithm/compose_voxel_features}
\input{algorithm/de-voxelization}

%% file: algorithm/voxelization.tex
\begin{listing}[!ht]
\begin{minted}[frame=single, breaklines, breakanywhere, fontsize=\scriptsize]{python}
from typing import List, Tuple
from typing_extensions import Literal

# Global variable indicating the shape to be sampled
SAMPLE_SHAPE: Literal["volume", "tri-plane", "tri-line", "center"] = "center"

def voxelization(gaussians, grid_size):
    """
    Convert a set of 3D Gaussians into a set of unique voxels and their associated features.
    This function does the following:
      1. Counts how many voxels will be needed for each Gaussian based on the current
         SAMPLE_SHAPE (e.g., "volume", "tri-plane", etc.).
      2. Generates voxel coordinates and features for all Gaussians.
      3. Removes duplicates, resulting in a unique set of voxel coordinates, their
         corresponding features, and an index map that indicates how each original
         voxel maps to the new unique list.
    Args:
        gaussians : List[dict]
            A list of 3D Gaussian parameters. Each Gaussian is typically a dictionary
            containing at least a "scale" key (and possibly position or other attributes)
            needed by the downstream functions.
        grid_size : float
            A hyperparameter defining the size of each voxel.
    Returns:
        uni_voxels_xyz : List[Tuple[float, float, float]]
            The (x, y, z) coordinates of the unique voxels.
        uni_voxel_feats : List
            The feature vectors corresponding to each unique voxel.
        inverse_indices : List[int]
            A list of indices that maps each original voxel back to its corresponding
            index in the unique voxel list.
        num_voxels_per_g: List[int]
            A list of number of voxels that are sampled from each 3D Gaussian.
    Notes:
        - The function relies on three helper functions which are assumed to be defined:
            1) `count_voxels_per_g(gaussians, grid_size)`
               Determines the number of voxels for each Gaussian given the sampling shape.
            2) `compose_voxel_features(gaussians, num_voxels_per_g)`
               Produces the (x, y, z) coordinates and any features for each voxel.
            3) `return_unique_voxels(voxels_xyz, voxel_feats)`
               Removes duplicate voxel coordinates and returns a list of unique
               coordinates, their combined features, and an inverse index array.
        - If you are implementing this in a parallel computing environment (e.g., CUDA),
          each step may be adapted for batch or GPU-based processing.
    """
    # 1. Determine how many voxels are required for each Gaussian.
    num_voxels_per_g = count_voxels_per_g(gaussians, grid_size)

    # 2. Generate voxel coordinates (xyz) and their associated feature vectors.
    #    These features might include Gaussian-related attributes or other data.
    voxels_xyz, voxel_feats = compose_voxel_features(gaussians, num_voxels_per_g)

    # 3. Deduplicate the generated voxels:
    #    - Get unique voxel coordinates,
    #    - Aggregate/merge features for any duplicates,
    #    - Obtain the inverse index mapping from original voxels to unique.
    uni_voxels_xyz, uni_voxel_feats, inverse_indices = return_unique_voxels(voxels_xyz, voxel_feats)

    return uni_voxels_xyz, uni_voxel_feats, inverse_indices, num_voxels_per_g
\end{minted}
\caption{Our voxelization process for 3D Gaussians.}
\label{code:voxelization}
\end{listing}

%% file: algorithm/compose_voxel_features.tex
\begin{listing}[!ht]
\begin{minted}[frame=single, breaklines, breakanywhere, fontsize=\scriptsize]{python}
from torch import Tensor

def compose_voxel_features(gaussians, num_voxels_per_g, grid_size):
    """
    Generate per-voxel feature vectors for a set of 3D Gaussians.
    This function iterates through each Gaussian and computes:
      1. The xyz location of each voxel in the Gaussian.
      2. A Mahalanobis distance-based influence (voxel opacity).
      3. Additional attributes such as RGB.
    Args:
        gaussians : List[dict]
            A list of Gaussian definitions
        num_voxels_per_g : List[int] or torch.Tensor
            Number of voxels associated with each Gaussian (e.g., from
            `count_voxels_per_g` or another precomputation).
        grid_size : float
            The size of each voxel cell, used to help determine voxel coordinates.
    Returns:
        voxel_features : List[torch.Tensor]
            A list of per-voxel feature tensors having [voxel_xyz, voxel_rgb, voxel_opacity].
    Notes:
        The function `get_voxel_location(...)` is assumed to provide the precise xyz coordinate of each voxel. Implementation details are omitted for readibility.
    """

    voxel_features = []
    voxel_xyzs = []

    # Iterate over Gaussians
    for g_idx, gaussian in enumerate(gaussians):
        # Retrieve the number of voxels expected for this Gaussian
        num_voxels = num_voxels_per_g[g_idx]

        # Extract Gaussian parameters for clarity
        mu = gaussian["mean"]  # Center location of a 3D Gaussian
        inv_cov = gaussian["inverse_3d_covariance_matrix"]
        opacity_factor = gaussian["opacity"]
        color_rgb = gaussian["rgb"]

        # Generate features for each voxel in this Gaussian
        for voxel_idx in range(num_voxels):
            # 1. Get the 3D coordinate for the current voxel
            voxel_xyz = get_voxel_location(voxel_idx, gaussian, grid_size)
            voxel_xyzs.append(voxel_xyz)

            # 2. Compute the Mahalanobis distance, Eq. 1 of the main manuscript
            dist = (voxel_xyz - mu).T @ inv_cov @ (voxel_xyz - mu)

            # 3. Compute voxel opacity using a Gaussian attenuation factor
            voxel_opacity = opacity_factor * torch.exp(-0.5 * dist)

            # 4. Concatenate features
            feature_vec = torch.cat([voxel_xyz, color_rgb, voxel_opacity], dim=1)
            voxel_features.append(feature_vec)

    return voxel_xyzs, voxel_features
\end{minted}
\caption{Pseudocode for the compose-voxel-features function.}
\label{code:compose_voxel_features}
\end{listing}

%% file: algorithm/de-voxelization.tex
\begin{listing}[!ht]
\begin{minted}[frame=single, breaklines, breakanywhere, fontsize=\scriptsize]{python}
from typing_extensions import Literal
from torch_scatter import scatter, segment_csr

# Global reduction mode (could also be passed as a parameter)
REDUCE: Literal["mean", "max"] = "max"

def de_voxelization(uni_voxels_pred, inverse_indices, num_voxels_per_g):
    """
    Aggregate network predictions from unique voxels back to full voxel predictions,
    and then further reduce them into per Gaussian predictions.
    This function performs two main steps:
      1. Unique Voxels -> All Voxels: Uses `scatter` to expand the predictions
         from the unique-voxel level back to the original list of all voxels, by
         indexing through `inverse_indices`.
      2. All Voxels -> Gaussians: Uses `segment_csr` to combine per-voxel
         predictions for each Gaussian. This is done by summing the number of
         voxels per Gaussian in `num_voxels_per_g`, which creates offsets used by
         `segment_csr` to group voxel predictions into Gaussian predictions.
    Args:
        uni_voxels_pred : torch.Tensor
            A tensor containing the predicted values (e.g., features or logits) for each
            unique voxel. Its shape could be (N_unique, C), where N_unique is the number
            of unique voxels and C is the dimensionality of the prediction (e.g. channels).
        inverse_indices : torch.Tensor
            A 1D tensor of indices mapping every original voxel to its corresponding
            index in the unique voxel array.
        num_voxels_per_g : torch.Tensor
            A 1D tensor indicating how many voxels belong to each Gaussian.
    Returns:
        gs_pred : torch.Tensor
            A tensor containing the final predictions at the Gaussian level, with shape
            (G, C).
    Notes:
        - If your pipeline needs a different reduction mode (e.g. `"sum"`) or you want
          to pass it in as a parameter, replace `REDUCE` accordingly.
    """

    # 1. Expand predictions from unique voxels to all voxels using `scatter`.
    #    - 'uni_voxels_pred' has shape (N_unique, C).
    #    - 'inverse_indices' has shape (M, ) with M >= N_unique (the total number of voxels).
    #    - 'scatter' will produce a new tensor of shape (M, C), where each entry is
    #      the value from the appropriate unique voxel. The method of combination
    #      (mean, max, etc.) is determined by `REDUCE`.
    voxels_pred = scatter(uni_voxels_pred, index=inverse_indices, reduce=REDUCE)

    # 2. Combine per-voxel predictions into Gaussian-level predictions using `segment_csr`.
    #    - We first compute offsets with cumulative sums of the number of voxels in each Gaussian.
    #    - The 'segment_csr' function then takes the voxel predictions and segments them
    #      into groups corresponding to each Gaussian, reducing via the same `REDUCE` rule.
    offset = torch.cumsum(num_voxels_per_g, dim=0)
    gs_pred = segment_csr(voxels_pred, ptr=offset, reduce=REDUCE)

    return gs_pred
\end{minted}
\caption{Our de-voxelization process for 3D Gaussians.}
\label{code:de_voxelization}
\end{listing}